\documentclass{article}

\usepackage[final]{neurips_2026_vericode}

\usepackage[utf8]{inputenc} 
\usepackage[T1]{fontenc}    
\usepackage{hyperref}       
\usepackage{url}            
\usepackage{booktabs}       
\usepackage{amsfonts}       
\usepackage{nicefrac}       
\usepackage{microtype}      
\usepackage{xcolor}         
 \usepackage{graphicx} 

\usepackage{amsmath}
 
\title{Repair, Retry, and the Cost of Hindsight: Verifier Feedback in Formal Theorem Proving}

\author{%
  Jack Li\thanks{Equal contributions, order determined by coin flip} \\
  Department of Computer Science\\
  Stanford University\\
  \texttt{jkli@stanford.edu} \\
   \And
  Sophia Li$^*$ \\
  Department of Computer Science \\
  Stanford University \\
  \texttt{sophyli@stanford.edu} \\
}

\begin{document}

\maketitle

\begin{abstract}

Verifier-guided self-repair conditions a model on feedback from failed attempts before it tries again, without updating model weights. Yet two findings in prior work are in tension: verifier feedback can improve repair of failed outputs, but self-repair itself can provide little-to-no advantage over independent resampling. Here, we provide an empirical reconciliation in Lean theorem proving through controlled one-step experiments that deconstruct repair by holding the preceding failure fixed while varying whether the failed proof is shown, whether the compiler error corresponds to that failure, and how the model is instructed to try again. We test frozen Qwen3.6-27B, a general-purpose model, and DeepSeek-Prover-V2-7B, a specialized prover model, on difficult MiniF2F problems. We find two opposing effects. Displaying the model's failed predecessor reduces success and sharply increases copying of it, while on Qwen adding that attempt's compiler error recovers the loss, to the level of the diagnostic shown alone. Replacing the predecessor with another theorem's failed proof largely removes the degradation on both models; on Qwen, replacing the diagnostic with another problem's error substantially reduces the recovery. Without the failed proof, correct diagnostics show no resolved advantage over wrong-diagnostic controls on either model. The failed-proof penalty persists with and without the retry instruction, whereas Qwen's diagnostic recovery is weaker without it. The predecessor penalty is accompanied by sharply increased copying of the model's own failed proof. Separate 11-problem sequential experiments on both models extend the comparison through 64 attempts: Qwen retains a correct-diagnostic benefit within failed-proof histories, whereas DeepSeek-Prover shows no resolved within-repair recovery. However, these recovery effects do not imply an advantage for self-repair as a test-time technique. In particular, no tested one-step feedback condition establishes an advantage over instruction-only resampling, nor does 64-attempt sequential repair with failed proofs and correct diagnostics at equal token budgets. These findings support interpreting the usefulness of verifier feedback separately from the net performance of self-repair, since a diagnostic can improve repair without overcoming the performance loss from conditioning on a failed attempt.

\end{abstract}

\section{Introduction}

Verifier-guided self-repair asks a language model to correct a failed attempt using feedback. Diagnostics can identify what went wrong, yet their apparent value depends on the comparison: proof-repair studies find them crucial \citep{first2023baldurwholeproofgenerationrepair}, while budget-aware evaluations find modest gains that independent resampling can match \citep{olausson2024selfrepairsilverbulletcode}. Repair prompts bundle a retry instruction, a failure notice, a diagnostic, and the failed attempt. We separate these components to ask whether diagnostics help repair an exposed failure and whether the resulting repair improves over starting fresh.

We study prompt-based feedback on frozen Qwen3.6-27B and DeepSeek-Prover-V2-7B models in Lean theorem proving. Unlike success on finite software tests, an accepted Lean proof establishes a fixed formal statement relative to the environment's definitions and trusted axioms, enabling evaluation without a learned judge. Our controlled one-step experiments hold failed predecessors fixed while varying the retry instruction, failed proof, diagnostic, and replacement context. Sequential experiments on separate 11-problem banks for both models then follow evolving repair histories through 64 attempts.

Our controls extend prior decompositions of repair feedback by separately testing predecessor correspondence and instruction-dependent diagnostic effects. Failed-code replacement asks whether harm is specific to the model's own predecessor rather than generic proof-like context; correct, ordinary-shuffled, and class/length-matched diagnostics test whether recovery depends on correspondence; and instruction crossings test whether that recovery is a property of the diagnostic alone. The fixed-predecessor design isolates these comparisons more directly than an evolving repair loop, while the sequential design tests whether their patterns persist when attempts become history.

Showing each model its own failed proof reduces success and increases predecessor copying, while replacing it with another theorem's failed proof restores performance toward the failure-only scaffold and nearly eliminates copying of the model's own predecessor. On Qwen, adding the correct diagnostic recovers this loss (40 problems, against 41 for the diagnostic alone), mostly on predecessors written in obsolete Lean~3 syntax (an exploratory subgroup), where it also beats a class/length-matched wrong diagnostic; no analogous within-repair recovery is resolved on DeepSeek-Prover. Sequentially, Qwen retains a resolved correct-versus-wrong diagnostic benefit within repair; DeepSeek does not, although its matched difference-of-effects
is nominally positive because matched wrong diagnostics perform better in the code-free protocol. No tested feedback condition establishes an advantage over instruction-only retrying in the one-step experiments, and sequential repair remains below or does not resolve an advantage over memoryless retrying at 64 attempts or at equal token budgets. Together, these results separate feedback that helps repair an exposed failure from evidence that repair itself is preferable to starting fresh.

\section{Related Work}

Verifier feedback supports incremental theorem proving \citep{thakur2024incontextlearningagentformal}, whole-proof repair and refinement \citep{first2023baldurwholeproofgenerationrepair,lin2025goedelproverv2scalingformaltheorem,li2026compilecompressboostingformal,requena2026minimalagentautomatedtheorem,wang2026learningrepairleanproofs}, and reflection, memory, or self-debugging \citep{shinn2023reflexionlanguageagentsverbal,chen2023teachinglargelanguagemodels}. Although intrinsic reasoning self-correction remains difficult \citep{huang2024largelanguagemodelsselfcorrect}, supplied error locations \citep{tyen2024llmsreasoningerrorscorrect} and richer type-error diagnostics \citep{krishnamurthi2026typeerrorablationaicoding} can improve correction. This literature includes trained repair systems; we study prompt-based feedback on frozen models.

Repair gains also need to be distinguished from the benefit of another attempt. Repeated sampling provides strong baselines \citep{brown2024largelanguagemonkeysscaling}, and once repair costs are counted, self-repair gains are modest and vary across benchmarks, difficulty levels, and models \citep{olausson2024selfrepairsilverbulletcode}. \citet{gu2026understandingperformancegapparallel} link sequential sampling losses to reduced exploration. In small code models, \citet{verma2026tryagaindontlook} report failed-output anchoring and no resolved execution-feedback advantage over a failed-code-plus-failure-notice control.

Placebo-controlled studies report feedback-content benefits within repair without establishing superiority over resampling \citep{iscan2026falsificationexposureinternallypreregistered}, whereas \citet{iscan2026formcontentpreregisteredplacebocontrolled} find no confirmed content advantage. Table~\ref{tab:prior_controls} compares the interventions most directly relevant to our controls. The rows are not claims of strict replication: domains, models, budgets, and matching targets differ.

\begin{table}[t]
\centering
\small
\begin{tabular}{p{0.18\linewidth}p{0.23\linewidth}p{0.23\linewidth}p{0.27\linewidth}}
\toprule
Study & Repair context & Feedback control & Additional separation \\
\midrule
\citet{verma2026tryagaindontlook} & Failed code under matched-budget retries & Placebo feedback; successful cross-task examples in a separate experiment & Failed-code exposure versus omission; cross-task examples supplement rather than replace the failed code \\
\citet{iscan2026falsificationexposureinternallypreregistered} & Failed code in one repair step & Generic fact bullets matched to the executed-feedback format & Blind resampling with matched generation count \\
This work & Fixed failed Lean proof and evolving histories & Cross-problem diagnostics, ordinary or class/length matched & Own-proof omission or replacement and retry-instruction crossings \\
\bottomrule
\end{tabular}
\caption{Operational comparison of controls in closely related test-time repair studies. ``Matched'' refers to each study's stated control variables, not exact token identity across studies.}
\label{tab:prior_controls}
\end{table}

We extend this investigation in Lean by holding failed predecessors fixed and comparing the effect of replacing correct diagnostics with other problems' diagnostics, both with and without failed proofs. Additional controls replace the predecessor with another theorem's failed proof and vary the retry instruction. Together with sequential evaluation, these comparisons characterize predecessor-specific harm, diagnostic correspondence, and prompt sensitivity separately from the complete repair procedure's performance against fresh retrying.

\section{Experimental setup}
\label{sec:setup}

We evaluate frozen Qwen3.6-27B (thinking disabled) on 167 difficult MiniF2F-test problems and DeepSeek-Prover-V2-7B on 137 of these, using model-specific failed predecessors; screening and bank construction are detailed in Appendix~\ref{sec:methods_setting}. In the sections and figures, ``code'' refers to the failed proof attempt and ``diagnostic'' to compiler error. Thirty problems in both banks write sums in a notation our verifier’s Mathlib no longer parses, so every condition fails on them; however, every one-step conclusion holds without them (Appendix~\ref{app:parseable}). Success requires Lean verification without errors or admitted proofs (Appendix~\ref{sec:methods_verification}). One-step experiments fix each predecessor and sample four independent continuations per condition, varying the instruction ``Produce a correct, complete proof.'' ($I$), failure notice (``You previously attempted to prove this theorem and it did not verify.''), proof exposure (failed solution attempts), and diagnostics (compiler errors) (Table~\ref{tab:prompt_components}; Appendix~\ref{sec:methods_onestep}). We also test model-generated reflection on Qwen (Appendix~\ref{app:framing_results}, Tables~\ref{tab:qwen} and~\ref{tab:noi}). Diagnostic controls replace the error with another problem's error, either unmatched or matched by coarse error class and length, with and without the failed proof. Appendix~\ref{sec:methods_relevance} details the unmatched diagnostic control and additional sampling for the Qwen comparison without failed proof present. A separate control replaces the failed proof with another theorem's failed proof.

Throughout, letters denote retained prompt content: $C$ is code from a previous failed attempt and $E$ is its corresponding compiler error. A subscript $s$ denotes ordinary cross-problem shuffling, $m$ denotes a wrong item selected by coarse error class and approximate length matching, and $o$ denotes another theorem's failed code. In the one-step study, $F_X$ denotes independent continuations from one fixed instruction-plus-failure-notice scaffold containing $X$; in the sequential study, $X$ denotes an evolving protocol that retains that content from preceding failed attempts. Thus, for example, $F_{CE}$ uses one fixed failed proof and error, whereas $CE$ accumulates failed proofs and errors along each trajectory. Table~\ref{tab:aliases} maps these names to the artifact's labels.

\begin{table}[ht]
\centering
\small
\begin{tabular}{lcccc}
\toprule
Condition & Instruction & Failure notice & Failed proof & Added feedback \\
\midrule
Bare & -- & -- & -- & -- \\
$I$ & Yes & -- & -- & -- \\
$F$ & Yes & Yes & -- & -- \\
$F_C$ & Yes & Yes & Yes & -- \\
$F_E$ & Yes & Yes & -- & Diagnostic \\
$F_{CE}$ & Yes & Yes & Yes & Diagnostic \\
\bottomrule
\end{tabular}
\caption{Central one-step prompt components. Every condition includes the theorem-completion task.}
\label{tab:prompt_components}
\end{table}

Sequential experiments evaluate each model on its own 11-problem bank, with ten trajectories per adaptive condition and up to 64 attempts. The seven-arm comparison includes failed-proof history alone ($C$), with corresponding ($CE$), ordinarily shuffled ($CE_s$), or class/length-matched wrong ($CE_m$) diagnostics, and corresponding or matched-wrong diagnostic histories without failed proofs ($E$ and $E_m$). Here, a corresponding diagnostic was produced by the same problem and attempt, whereas a wrong diagnostic comes from another problem. Each history is a sliding window: once the displayed history exceeds about 24{,}500 characters, whole attempts are dropped, oldest first. Late Qwen $CE$ prompts therefore show roughly the last seven failed attempts, and DeepSeek-Prover’s, whose attempts are shorter, about 18. DeepSeek-Prover's prompts place the Lean preamble inside the theorem's code fence, its native format; Qwen's show it only inside displayed failed proofs (Appendix~\ref{app:prompts}). Memoryless instruction-only retrying ($I$, prompted ``Now produce a correct, complete proof.'') is estimated by pooling 640 draws per problem. Qwen failure-marker and random-word protocols are reported separately in the appendix. Comparisons are within model and do not isolate a causal effect of model choice (Appendix~\ref{sec:methods_sequential}).

Pass@$k$ measures the probability of at least one verified proof within $k$ attempts, averaged across problems. We report paired percentage-point differences with confidence intervals (Appendix~\ref{sec:methods_analysis}). Primary budgets count attempts; Appendix~\ref{app:token_results} repeats the sequential comparison at equal token budgets.

\section{Results}
\label{sec:results}
\begin{figure}[t]
\centering
\includegraphics[width=1 \linewidth]{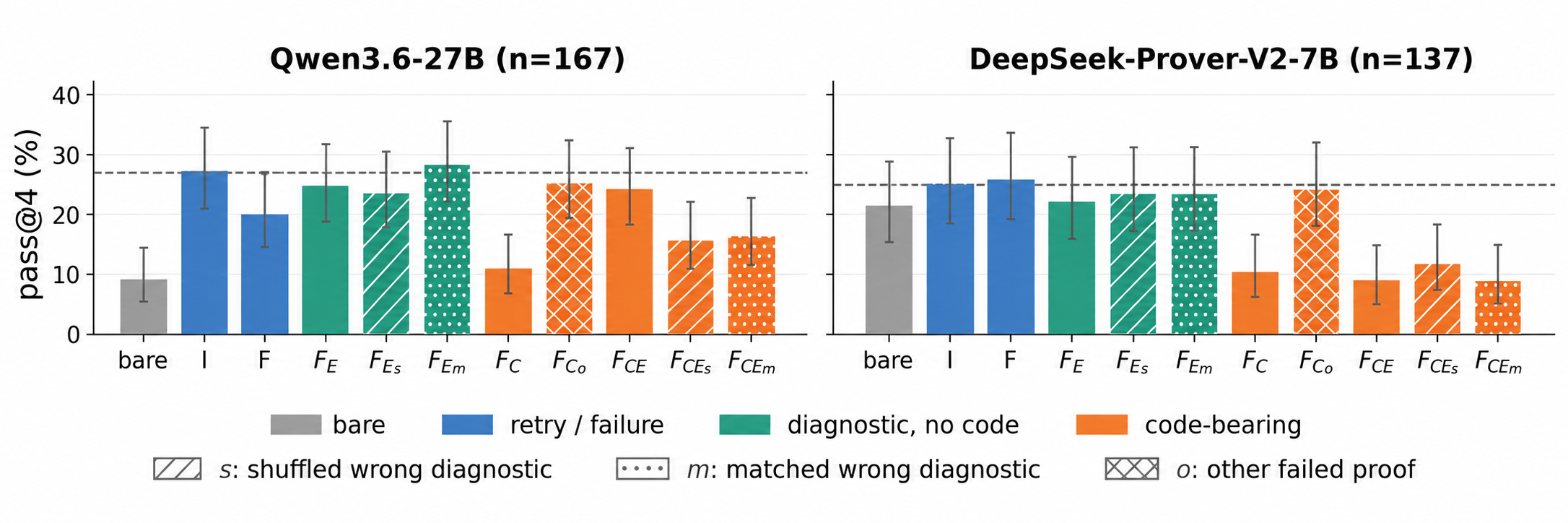}
\caption{Controlled one-step pass@4. Dashed lines mark instruction-only performance. $s$: shuffled wrong diagnostic; $m$: class/length-matched wrong diagnostic; $o$: another theorem's failed proof. Whiskers are marginal 95\% problem-bootstrap intervals; paired contrast intervals are reported in text.}
\label{fig:decomposition}
\label{fig:copy}
\end{figure}

Figure~\ref{fig:decomposition} summarizes the one-step controls; Fig.~\ref{fig:sequential} shows sequential comparisons. Full ablations, precision analyses, and sequential details are in Appendix~\ref{app:results_details}. The net repair comparison motivates the component analyses that follow: diagnostic correspondence, failed-predecessor exposure, and the retry instruction. 

\subsection{Repair versus retry}
\label{sec:retry_results}
Qwen solves 45/167 problems with instruction-only retrying versus 40 with failed attempt plus the correct diagnostic ($F_{CE}$); DeepSeek-Prover solves 34/137 versus 12. Thus, the controlled one-step experiments do not establish an advantage for these repair conditions over independent resampling. Per sample, Qwen's $F_{CE}$ is level with $I$ ($-0.6$~pp; $[-4.5,+3.3]$), so its pass@4 gap reflects passes spread over fewer problems rather than fewer passes; no feedback condition shows a resolved per-sample advantage on either model (Appendix~\ref{app:precision_results}). The preregistered sequential experiment on 11 Qwen problems (10 trajectories per problem in each arm) also does not establish an advantage for repair. At $k=64$ sequential attempts, code plus the correct diagnostic ($CE$) reaches 44.5\% versus 62.8\% for memoryless instruction-only retrying ($I$), a difference of $-18.3$ pp (95\% paired problem-bootstrap interval $[-41.8,+5.3]$); dropping any one problem never makes this estimate positive (range $-25.1$ to $-12.1$~pp; Appendix~\ref{app:sequential_results}). Failed-attempt history alone ($C$) underperforms retrying ($-40.1$ pp; $[-62.2,-17.0]$).

For the DeepSeek-Prover sequential experiment at $k=64$, on its own 11-problem bank with ten trajectories per problem, $CE$ and $C$ each solve 27/110 trajectories (24.5\%), versus 75.3\% for memoryless $I$. DeepSeek's $CE-I$ and $C-I$ are both $-50.7$ pp, with 95\% paired problem-bootstrap intervals $[-66.4,-34.5]$ and $[-67.6,-33.6]$, respectively.

Code-free diagnostic histories also show no resolved advantage over memoryless retrying (Qwen $E-I=+0.8$~pp, $[-13.2,+16.4]$; DeepSeek $E-I=-8.0$~pp, $[-26.2,+6.5]$). DeepSeek's matched-wrong-diagnostic history has the highest $k=64$ point estimate ($E_m-I=+6.5$~pp, $[-3.4,+18.3]$), which we do not read as a feedback benefit.

Adaptive arms start from the bare prompt, since no history exists at attempt 1. Conditioning on failure of attempt 1 leaves $CE-I$ at least as unfavorable to repair ($-19.1$~pp, $[-42.3,+4.5]$, on Qwen; $-56.5$~pp, $[-72.5,-39.5]$, on DeepSeek-Prover), though it does not equalize the first prompts (Appendix~\ref{sec:methods_sequential}).

Matching attempts does not match cost. A sequential repair prompt carries its history, so a $CE$ attempt averages about 9{,}100 input tokens on Qwen and 9{,}700 on DeepSeek-Prover, against 115 and 146 for a memoryless draw. Giving both protocols the same cumulative token cap per trajectory, each stopping at its first verified proof, does not change the conclusion (Appendix~\ref{app:token_results}). At 200{,}000 total tokens, $CE$ trails memoryless retrying by $25.9$~pp on Qwen (bootstrap interval $[-50.4,-2.4]$, though an exact randomization test does not resolve it, $p=0.066$) and by $73.5$~pp on DeepSeek-Prover ($[-89.8,-53.5]$; $p=0.002$). Counting only generated tokens, the accounting most favorable to repair, Qwen's $CE$ is level with memoryless retrying at 20{,}000 tokens ($-0.4$~pp; $[-19.2,+17.7]$), while DeepSeek-Prover's still trails ($-28.2$~pp at 5{,}000 tokens; $[-43.0,-15.2]$). Under none of these accountings does repair gain an advantage.

\subsection{What the diagnostic contributes}
\label{sec:diagnostic_results}
In the one-step experiments without the failed proof, correct diagnostics show no resolved advantage over either wrong-diagnostic control on either model (Qwen correct/shuffled/matched-wrong: 41/39/47; DeepSeek-Prover: 30/32/32). A higher-precision Qwen comparison pooling 16 additional attempts per condition estimates correct-minus-shuffled at $+1.73$ pp, with a problem-bootstrap 95\% interval of $[-2.06,+5.62]$; fixed-bank equivalence is reported separately in Appendix~\ref{app:precision_results}.

With Qwen's own failed proof present, the correct diagnostic $F_{CE}$ solves 40 problems versus 18 with code alone $F_C$, recovering $+13.2$ pp (95\% paired-bootstrap interval $[+7.2,+19.2]$). It also exceeds the shuffled $F_{CE_s}$ and class/length-matched $F_{CE_m}$ wrong diagnostics (26 and 27 problems); correct versus matched gains $+7.8$ pp ($[+1.8,+13.8]$), while the two wrong controls show no resolved difference ($+0.6$ pp; $[-4.8,+6.0]$). This recovery is concentrated in the 50 problems whose frozen predecessor is written in Lean~3 syntax, which a Lean~4 verifier rejects (an exploratory subgroup; 48 of the 68 whose diagnostic names an unknown identifier; Appendix~\ref{app:behavior_results}). There, $F_{CE}$ solves 25 problems versus 8 for $F_C$ ($+34.0$~pp; $[+20.0,+48.0]$) and 15 for $F_{CE_m}$ ($+20.0$~pp; $[+6.0,+34.0]$); on the other 117 the same contrasts are $+4.3$~pp ($[-0.9,+9.4]$) and $+2.6$~pp ($[-3.4,+8.5]$). Thus, Qwen's one-step recovery mainly reflects diagnostics that flag the predecessor's obsolete syntax. The exploratory matched interaction $(F_{CE}-F_{CE_m})-(F_E-F_{E_m})$ is $+11.38$~pp ($[+2.40,+19.76]$); like $F_{CE}-F_{CE_m}$, it is significant after Benjamini--Hochberg but not Holm adjustment across the one-step exploratory comparisons (Appendix~\ref{app:stats}). It combines the resolved within-repair contrast (13 of its 19 problems) with a small, unresolved reversal without code ($F_E-F_{E_m}=-3.6$~pp). DeepSeek-Prover shows no analogous rescue ($F_{CE}$, $F_{CE_s}$, and $F_{CE_m}$ solve 12, 16, and 12 problems, versus 14 for $F_C$). On 37 of its problems the matched donor's text equals the correct diagnostic, but excluding them leaves $F_{CE}-F_{CE_m}$ at $0.0$~pp ($[-5.0,+5.0]$; Appendix~\ref{app:ext3}). The Qwen ordinary-shuffle code-by-correctness interaction is positive but unresolved ($+7.2$ pp; $[-1.2,+15.6]$).

For Qwen, the sequential experiment strengthens this result. At $k=64$, adding the correct diagnostic to failed-code history improves success over code alone ($CE-C=+21.8$ pp; 95\% paired problem-bootstrap interval $[+10.9,+32.7]$), over an ordinarily shuffled wrong diagnostic ($CE-CE_s=+18.2$ pp; $[+8.2,+28.2]$), and over a class/length-matched wrong diagnostic ($CE-CE_m=+23.6$ pp; $[+12.7,+35.5]$). Exact paired randomization tests agree ($p=0.012$, $0.012$, and $0.004$; Appendix~\ref{app:sequential_results}). Without code, the matched correct-versus-wrong diagnostic contrast is smaller and unresolved ($E-E_m=+6.4$ pp; $[-6.4,+19.1]$). Qwen's $E$ arm was generated in an earlier launch than $E_m$ and the code-bearing arms, so contrasts involving it also carry launch-to-launch variation, which a replicate memoryless pool puts at 4.2~pp at $k=64$ (Appendix~\ref{app:sequential_results}). The preregistered ordinary-shuffle interaction $(CE-CE_s)-(E-E_m)$ is positive but unresolved at $+11.8$ pp ($[-8.2,+29.1]$) and uses different wrong-diagnostic policies on its two sides. The separately preregistered matched interaction $(CE-CE_m)-(E-E_m)$ is $+17.3$ pp ($[-1.8,+34.5]$), also unresolved for Qwen. Thus, corresponding diagnostic content has a resolved benefit within Qwen's code-bearing repair histories, without resolving either interaction at the 64-attempt endpoint.

DeepSeek-Prover does not reproduce this within-repair recovery at $k=64$: $CE-C=0.0$ pp ($[-3.6,+3.6]$), $CE-CE_s=+1.8$ pp ($[-10.0,+13.6]$), and $CE-CE_m=+1.8$ pp ($[-5.5,+9.1]$), with both wrong-diagnostic arms solving 25/110 trajectories. Without code, corresponding-diagnostic history solves 74/110 trajectories versus 90/110 for matched-wrong-diagnostic history ($E-E_m=-14.5$ pp; $[-30.0,-2.7]$; exact randomization $p=0.063$). The ordinary-shuffle difference-of-effects is unresolved ($+16.4$~pp; $[-1.8,+34.5]$). The matched interaction excludes zero under the registered bootstrap ($+16.4$~pp; $[+0.9,+35.5]$) but not under an exact randomization test ($p=0.11$), and its sign comes from the negative code-free contrast ($E-E_m=-14.5$~pp); the within-repair component $CE-CE_m$ is $+1.8$~pp and unresolved. The absence of a resolved within-repair benefit on DeepSeek is not evidence of equivalence, and with different banks, prompts, and toolchain fit (Section~\ref{sec:limitations}) the contrast between models is not causal.

\subsection{The role of the failed predecessor}
\label{sec:feedback}
Starting from the instruction-plus-failure-notice scaffold $F$ (33 Qwen solves; 35 on DeepSeek-Prover), adding the model's own failed proof lowers success to 18 and 14 ($F_C-F=-9.0$~pp, $[-15.6,-2.4]$, preregistered with Holm-adjusted $p=0.055$; and $-15.3$~pp, $[-21.9,-9.5]$). Against instruction-only retrying the deficit is $-16.2$~pp ($[-22.8,-9.6]$) and $-14.6$~pp ($[-20.4,-8.8]$); on Qwen part of it comes from $F$ itself, which trails $I$ by $7.2$~pp ($[-12.6,-2.4]$). On Qwen this one-step penalty is concentrated in the 50 Lean~3 predecessors ($-26.0$~pp; $[-40.0,-12.0]$), where exposure leads the model to reproduce Lean~3 syntax; on the other 117, where success is near the floor ($F$ solves 12), it is $-1.7$~pp ($[-8.5,+5.1]$), although exact copying is as frequent (15.8\% versus 18.5\% of samples). DeepSeek-Prover's predecessors are its own Lean~4 proofs, and its penalty is not concentrated in any such subset (Appendix~\ref{app:behavior_results}). Replacing that predecessor with another theorem's failed proof improves success by $+14.4$ pp on Qwen (42 versus 18; 95\% interval $[+7.8,+21.6]$; again mostly on Lean~3 predecessors) and $+13.9$ pp on DeepSeek-Prover (33 versus 14; $[+8.0,+20.4]$). Relative to $F$, the unrelated-proof condition solves 42 versus 33 problems on Qwen and 33 versus 35 on DeepSeek-Prover; both 95\% confidence intervals for these differences include zero. Own-predecessor copying simultaneously falls from 16.6\% to 0.6\% on Qwen and from 60.8\% to 1.6\% on DeepSeek-Prover (failed-predecessor vs unrelated failed code). Near-copying (sequence similarity at least 0.9) shows the same pattern: relative to $F$, exposure raises it by $21.0$~pp on Qwen ($[+15.9,+26.2]$) and $65.7$~pp on DeepSeek-Prover ($[+58.9,+72.1]$), and replacement removes the increase (Appendix~\ref{app:behavior_results}). This supports predecessor-specific harm in the tested comparison over generic proof-context explanation.

The sequential controls reproduce failed-history harm in both tested banks. Under the same class/length-matching rule, $CE_m-E_m$ is $-36.4$ pp on Qwen ($[-50.0,-21.8]$) and $-59.1$ pp on DeepSeek-Prover ($[-71.8,-44.5]$). This exploratory comparison uses different donor pools and generated histories, so it compares evolving protocols rather than isolating the effect of adding one fixed failed proof. The ordinary-shuffle comparison, $CE_s-E_m$, is $-30.9$ pp on Qwen ($[-49.1,-13.6]$) and $-59.1$ pp on DeepSeek-Prover ($[-76.4,-40.0]$); it is reported separately because its code-bearing and code-free sides use different wrong-diagnostic policies.

\subsection{The role of the retry instruction.}
\label{sec:instruction_results}
We experiment with removing the retry instruction while retaining the theorem-completion task, failure notice, and any proof or diagnostic blocks. In these one-step ablations, adding the failed proof reduces solved problems from 46 to 23 on Qwen and from 32 to 8 on DeepSeek-Prover, so the failed-proof penalty remains visible without the retry instruction. Adding the correct diagnostic beside the failed proof raises these counts to 28 and 11, respectively. Qwen's observed diagnostic gain is smaller without the retry instruction (5 additional problems) than with it (22; 18 to 40); the direct exploratory interaction is $+10.18$ pp ($[+2.40,+17.96]$). This suggests instruction dependence for the measured Qwen recovery contrast (the interaction is significant after Benjamini--Hochberg but not Holm adjustment; Appendix~\ref{app:stats}), while the exposure penalty itself persists without the instruction. Full comparisons appear in Appendix~\ref{app:framing_results}, Tables~\ref{tab:noi} and~\ref{tab:prover}, and Fig.~\ref{fig:framing_details}.

\begin{figure}[t]
\centering
\includegraphics[width=0.92\linewidth]{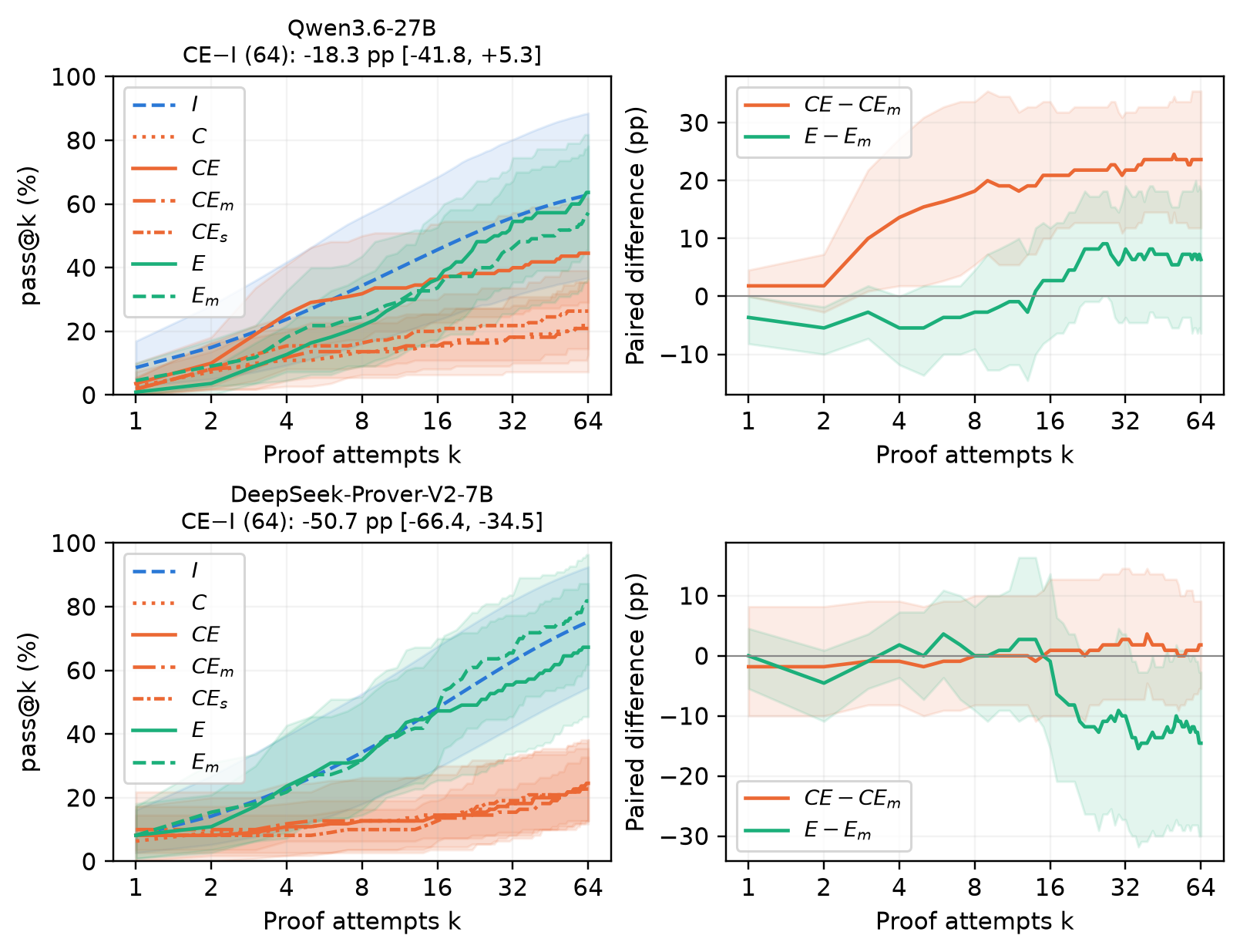}
\caption{\textbf{Sequential repair and diagnostic-content effects on both models.}
Qwen (top) and DeepSeek-Prover (bottom) each use their own 11-problem bank.
\textbf{Left:} success by attempt $k$ for memoryless instruction-only resampling ($I$), failed-proof history alone ($C$), failed-proof histories with correct ($CE$), ordinary-shuffled ($CE_s$), or matched-wrong ($CE_m$) diagnostics, and code-free diagnostic histories ($E,E_m$). Adaptive curves are fractions of 110 trajectories solved; $I$ is mean pooled pass@$k$ from 640 draws per problem.
\textbf{Right:} matched corresponding-minus-wrong diagnostic effects with failed proofs ($CE-CE_m$, orange) and without them ($E-E_m$, green). Their difference defines the registered matched interaction reported in the text and Appendix~\ref{app:sequential_results}.
Shading shows pointwise 95\% problem-bootstrap intervals: marginal for left-panel curves and paired for right-panel contrasts.}
\label{fig:sequential}
\end{figure}

\section{Discussion}
\label{sec:discussion}

A diagnostic can be useful for repairing a failed proof without making repair better than a new attempt. In our one-step conditions with the retry instruction, Qwen's correct diagnostics improve success over failed-proof exposure alone and over wrong diagnostics, yet the resulting repair condition does not establish an advantage over instruction-only resampling (Sections~\ref{sec:retry_results} and~\ref{sec:diagnostic_results}); DeepSeek-Prover shows no resolved within-repair diagnostic benefit. Without the failed proof, neither model shows a resolved advantage for correct over wrong diagnostics in these one-step experiments. A related pattern appears in the primary sample of \citet{iscan2026falsificationexposureinternallypreregistered}: execution feedback improves repair in small code models relative to bare failed code and format-matched generic bullets, while code plus feedback and blind resampling have equal pooled success counts. We extend this by fixing the failed proof and comparing correct diagnostics with other problems' errors, including class/length-matched ones; Qwen's advantage over these controls shows that correspondence to the failed proof contributes to recovery, chiefly where the diagnostic flags obsolete Lean~3 syntax.

Displaying the model's own failed proof reduces one-step success on both models, on Qwen mainly for Lean~3 predecessors (Section~\ref{sec:feedback}). Replacing it with another theorem's failed proof both improves success and sharply reduces own-predecessor copying, from 16.6\% to 0.6\% on Qwen and from 60.8\% to 1.6\% on DeepSeek-Prover. The gain in verified success under replacement argues against an explanation based solely on adding proof-like context. Unlike \citet{verma2026tryagaindontlook}'s successful cross-task examples, which add potentially useful successes beside the failed code, our replacement substitutes an unrelated failure for the predecessor, testing whether arbitrary failed-proof context reproduces the degradation. The improved success and reduced exact copying under predecessor replacement are consistent with accounts linking conditioning on prior outputs to repetition and reduced exploration \citep{verma2026tryagaindontlook,gu2026understandingperformancegapparallel}; Gu et al. measure broader embedding similarity and exploration, whereas our copying measures are exact and near recurrence of the predecessor (Appendix~\ref{app:behavior_results}). Neither establishes that copying causes the performance loss. Qwen's diagnostic recovery also qualifies the comparison with \citet{verma2026tryagaindontlook}'s experiments, which found no detectable feedback-content benefit: in our Qwen setting, conditioning on the failed proof is harmful, but its compiler error still improves repair.

The instruction ablations help distinguish the effects of failed-proof exposure from the instruction to try again and prove correctly. Removing the retry instruction to produce a complete and correct proof does not eliminate the failed-proof penalty, while the theorem-completion task and failure notice remain (Section~\ref{sec:instruction_results}). Qwen's diagnostic recovery is smaller without that instruction, and the exploratory interaction, which does not survive Holm adjustment, resolves a difference between the measured instruction conditions. Thus the recovery observed on Qwen depends partly on how the model is prompted to use the compiler error, whereas failed-proof exposure remains harmful without that instruction.

The sequential experiments extend the evidence beyond conditioning on a fixed failed predecessor (Fig.~\ref{fig:sequential}) and one step of repair. The continued disadvantage of failed-proof history on both banks shows that predecessor harm also occurs when the model can generate new attempts over successive rounds. Qwen's diagnostic recovery persists against both ordinary-shuffled and matched wrong diagnostics, but does not establish an advantage over fresh resampling at 64 attempts. DeepSeek-Prover shows no resolved correct-diagnostic benefit within repair. Its matched interaction is nominally positive because corresponding diagnostics underperform matched wrong diagnostics in the code-free protocol, not because correct diagnostics resolve recovery beside failed code, and it is not robust to an exact randomization test. Thus, a diagnostic benefit can persist across repeated repair without establishing that retaining failed attempts is a better use of the attempt budget.

A practical question raised by these results is what content to keep in context after a prover fails. Existing prover designs make different retention choices: COPRA, for example, conditions tactic selection on proof states and records of unsuccessful tactics \citep{thakur2024incontextlearningagentformal}, while \citet{li2026compilecompressboostingformal} correct errors locally from compiler output rather than accumulating a history of attempts. \citet{requena2026minimalagentautomatedtheorem} find that iterative refinement beats repeated single-shot generation in sample efficiency and cost; their agent pairs frontier models with library search and context management, whereas we condition frozen models on raw failed attempts, so the results need not conflict, though we do not test curated histories. For whole-proof repair, our controls motivate three comparisons: a fresh-attempt baseline to evaluate the net benefit of repair, corrupted diagnostics to test the contribution of corresponding feedback, and predecessor replacement or omission to assess the effect of the failed proof. For practitioners, the retention choice our results most directly support is to discard the failed proof, which carries the penalty on both models. Whether the diagnostic is then worth keeping we do not resolve: without the failed proof, correct and wrong diagnostics are indistinguishable on both models. On this evidence neither choice should be expected to beat resampling from a well-written instruction at the same attempt budget, and at equal total or input tokens the comparison favors resampling further still (Appendix~\ref{app:token_results}).

\section{Limitations}
\label{sec:limitations}

The one-step experiments condition on a single failed predecessor per theorem; failed proofs vary in length, error class, and proximity to a working proof, so a different predecessor could yield a different penalty or recovery. Attempt 2 of each sequential trajectory repairs a freshly sampled predecessor, and across about a hundred such predecessors per model the exposure-to-copying association replicates (Appendix~\ref{app:behavior_results}); attempt-2 success is too close to the floor to test the success penalty, which would need several fixed predecessors per theorem.

We study two frozen models, Qwen3.6-27B \citep{qwen3.6-27b} with its thinking mode disabled and DeepSeek-Prover-V2-7B \citep{ren2025deepseekproverv2advancingformalmathematical}, on model-specific banks of difficult MiniF2F problems \citep{zheng2022minif2fcrosssystembenchmarkformal}, with limited prompt formats and decoding settings. Each sequential study contains only 11 problems, so inference over problems is comparatively imprecise for some contrasts---most notably the preregistered $CE-I$ and difference-of-effects endpoints---and the depth results characterize these deliberately difficult, differently constructed banks rather than a broad MiniF2F population. The bank size was fixed by generation cost, and the median $k=64$ endpoint interval is 35.5~pp wide on Qwen and 29.1~pp on DeepSeek-Prover. One generalist and one specialist do not establish a general distinction between these model categories, and the two models differ in more than their banks. DeepSeek-Prover's prompts carry the Lean preamble in the theorem's fence while Qwen's do not, and within Qwen the preamble is visible only in code-bearing arms, so code-versus-code-free contrasts also vary its visibility (within-code contrasts and predecessor replacement do not). Toolchain drift is a further difference: we verify on Lean v4.29.0 and a 2026 Mathlib, whereas DeepSeek-Prover-V2 was evaluated on Lean 4.9 with an August 2024 Mathlib. Every identifier its predecessor diagnostics flag, and 93\% of the unknown identifiers in its failed one-step samples, is a name valid in that older library and since renamed, so its weak response to diagnostics may partly reflect not knowing the replacements; on Qwen, the one-step effects concentrate on Lean~3 predecessors (Appendix~\ref{app:drift}). Generalization to other models, difficulty levels, and domains remains untested.

Our primary budgets count proof attempts rather than equal token, latency, or compute expenditure. A repair prompt costs 66 to 80 times a memoryless prompt's input tokens, so attempt matching is generous to repair; the token-matched comparison (Appendix~\ref{app:token_results}) counts tokens rather than FLOPs or latency.

Diagnostic controls do not isolate semantic correctness perfectly. In the one-step study, we compare corresponding diagnostics with both unmatched wrong diagnostics and diagnostics matched by coarse error class and length; matching controls these properties but not specificity or every other feature of the message. Sequential $E_m$ and $CE_m$ use the same matching principle but draw from different regime-specific error pools. Sequential length matching is approximate, especially for DeepSeek-Prover (Appendix~\ref{sec:methods_sequential}). These controls reduce superficial diagnostic mismatch but do not eliminate it, and comparisons between evolving protocols do not isolate one fixed code-exposure intervention. The predecessor-replacement donor is an unmatched cross-problem draw; the effect holds on the small same-class and length-matched subsets (Appendix~\ref{app:ext3}), but this control alone does not separate correspondence from donor length and error class.

Finally, Qwen's sequential diagnostic-content difference-of-effects is not resolved at the preregistered $k=64$ endpoint, despite resolved $CE-CE_s$ and $CE-CE_m$ effects within code-bearing repair. DeepSeek's matched interaction excludes zero under the registered bootstrap but not under an exact randomization test, and its sign comes from the negative code-free contrast rather than resolved within-repair recovery. The two added one-step interactions are exploratory; they are significant after Benjamini--Hochberg but not Holm adjustment (Appendix~\ref{app:stats}). We therefore conclude that problem-specific diagnostic content helps Qwen within code-bearing repair, but do not infer a common mechanism across models. Likewise, the strong association between failed-proof exposure and predecessor repetition supports a behavioral anchoring account but does not establish repetition as the causal mediator of lower success. The experiments these limitations call for are generation runs outside the scope and timeline of this study: resampling padded to the repair prompt's length, counterfactual diagnostics that perturb only an identifier or a source location, and larger banks on further models and easier problem distributions.

\paragraph{Broader impacts.} This controlled study uses public models and a public benchmark, with no human subjects. Its findings bear on verifier-guided agents, including code-repair agents with security relevance, but cut against the capability-increasing reading: the retention choice we support, discarding the failed attempt, simplifies the loop and still does not beat resampling. Because Lean proofs are machine-checked, the results do not depend on a gameable learned judge, and a negative result about a common test-time technique can save practitioners compute.

\begin{ack}
Experiments used the FarmShare cluster, operated by Stanford Research Computing, and the Clariden cluster on Alps. This work was supported by a grant from the Swiss National Supercomputing Centre (CSCS) under project ID a0190 on Alps.  We are grateful to Jonas H\"{u}botter and Carlos Guestrin for their support and advice. Additionally, we thank Stanford Research Computing and CSCS for these resources and support, and the VeriCodeGen reviewers and area chair for their feedback.

\end{ack}

\bibliography{cite} 
\bibliographystyle{plainnat}

\clearpage

\appendix

\section{Methods}
\label{sec:methods}

\subsection{Task, models, and failure banks}
\label{sec:methods_setting}
We study simple prompt-based verifier feedback for Lean proof completion with frozen generalist Qwen3.6-27B and specialist DeepSeek-Prover-V2-7B models. Each task fixes the theorem statement and library environment; the model supplies a proof body for verification.

We restrict the Qwen bank to MiniF2F-test problems with estimated single-attempt success below 25\% in an initial screening pool. Excluding four framing-development problems leaves 167 problems, with a mean recorded screening rate of 2.35\%; 132 have no recorded screening successes. Each is paired with a fixed failed proof and diagnostic. For DeepSeek-Prover, we retain 137 of these problems with a usable failure among four initial samples and use that model's own failed proof. Only 44 of these 137 problems are solved under any DeepSeek-Prover one-step condition, so its one-step contrasts are carried by those 44. Thirty problems, in both banks, state sums or products as \texttt{$\sum$ x in s}, notation that the verifier's Mathlib (v4.29.0) no longer parses; because verification uses the dataset statement, all 6{,}360 attempts on them fail with the same parse error, in every condition. All 30 recorded zero screening successes, which is how the difficulty screen admitted them. Neither sequential bank contains any of them. Comparisons are paired within model-specific banks, not interpreted as a head-to-head model ranking.

\paragraph{Bank construction and its scope.} The upstream screening pool itself is not bundled, so the frozen manifest and the recorded exclusions, rather than a re-executable screening rule, are what reproduce the bank; the historical field name recording 256 draws does not certify a uniform 256-draw procedure. DeepSeek-Prover's 137 problems are those of the same manifest for which that model produced a usable failed predecessor, after excluding 30 problems it solved on every bare attempt. The two sequential banks are separate 11-problem sets. Qwen's is a frozen list recorded as a random draw of 12 problems from the 171-problem screened pool; the seed and draw script were not retained, so the list is reproduced as frozen rather than re-drawn, and one problem was later removed for a theorem-extraction artifact. The release records that the draw predates the four framing-development holdouts; all four are among its eleven retained problems, which an independent choice would make very unlikely (about 1 in 69{,}000), so the holdouts were evidently taken from the drawn list. DeepSeek-Prover's is selected from a 32-draw cold screen over its own 137 problems to approximately match Qwen's screening rates (Appendix~\ref{sec:methods_sequential}). Four of the eleven Qwen sequential problems are the same problems held out of the one-step bank for framing development; every sequential contrast is a within-bank comparison between arms sharing one scaffold, so this does not favor an arm, but the Qwen sequential bank is not independent of prompt development. These are populations selected to study repair after failure on hard problems, and the pass rates and penalties we report are conditional on them rather than representative of MiniF2F. Qwen's retained predecessors are shorter than the model's current failures on the same problems (mean within-problem length percentile 0.41) and over-represent the syntax class (51\% versus 36\%), consistent with their origin in an earlier screening run that produced many Lean~3-style proofs. DeepSeek-Prover's predecessors are typical of its failures (percentile 0.49; the predecessor has its problem's modal error class on 108 of the 135 problems with other failures on record). Appendix~\ref{app:behavior_results} reports how the Qwen effects vary with this composition.

\subsection{Controlled one-step experiments}
\label{sec:methods_onestep}
We hold the preceding failure fixed to isolate individual prompt components. For each problem and condition, we generate four independently sampled continuations of the same prompt. Outputs do not update subsequent prompts: this is one repair step sampled four times, not a four-step repair trajectory.

The bare prompt begins ``Complete the following Lean 4 code:'' followed by the theorem. The instruction-only condition ($I$) prepends ``Produce a correct, complete proof.'' and retains the entire bare task prompt. The failure scaffold ($F$) adds ``You previously attempted to prove this theorem and it did not verify'' while retaining that instruction. We then add the failed proof ($C$), its compiler error ($E$), both, or a model-written reflection ($R$). Reflection is tested on Qwen only: a separate generation receives the theorem, failed proof, and diagnostic and produces a short critique, frozen across the four continuations. Table~\ref{tab:prompt_components} summarizes the central conditions. Supplementary ablations remove the instruction or replace the failure sentence with alternative framing or random-word text; the failure sentence alone is therefore distinct from $F$.

Table~\ref{tab:prompt_components} in the main text summarizes the central components; the additional $F_R$ condition replaces the diagnostic with a frozen model-written reflection.

\subsection{Testing diagnostic relevance}
\label{sec:methods_relevance}
The ordinary-shuffle one-step comparisons distinguish problem-specific information from the generic effect of including verifier feedback. We replace the true diagnostic with one from another problem while retaining the diagnostic block's position and the surrounding prompt, testing this control both with and without the failed proof. Replacement diagnostics are sampled with replacement from the model's bank, must differ from the correct diagnostic text, and remain fixed across samples and code-presence conditions. These shuffled replacements are not matched for error type or length. The matched diagnostic controls use wrong diagnostics matched by coarse error class and length, with and without the failed proof ($F_{E_m}$ and $F_{CE_m}$). The predecessor-replacement control $F_{C_o}$ is not matched on any property: it substitutes the failed proof of the same seeded cross-problem partner that $F_{E_s}$ and $F_{CE_s}$ draw their diagnostic from, so the replacement proof differs from the model's own predecessor in length and error class as well as in correspondence to the theorem. All three use the same frozen banks and four samples per condition. On 37/137 DeepSeek-Prover problems, the matched wrong diagnostic is byte-identical to the correct diagnostic; this limitation and the independently shuffled comparison are reported in Appendix~\ref{app:ext3}.


For the code-free Qwen comparison, we collect 16 independent attempts per problem and condition in four batches. We pool these attempts to estimate success within four draws and report the batch-based estimate as a sensitivity check; the failed predecessor and assigned replacement diagnostic remain fixed.

\subsection{Sequential experiments}
\label{sec:methods_sequential}
We evaluate Qwen and DeepSeek-Prover separately, each on its own 11-problem bank with ten adaptive trajectories per condition and a maximum of 64 attempts. Every adaptive trajectory stops at its first verified proof or reaches the horizon. The banks share only \texttt{imo\_1962\_p2} and \texttt{mathd\_numbertheory\_314}; comparisons are paired within model and do not isolate a causal effect of model choice.

The Qwen bank is a frozen list recorded as a random draw of 12 problems from the 171-problem pool screened for estimated single-attempt success below 25\%; the draw's seed and script were not retained, so the list is reproduced as frozen rather than re-drawn. The threshold was an upper bound rather than a target difficulty: the eleven retained problems have recorded screening rates of 0 (four problems), 0.4, 0.8, 1.2, 1.6, 3.5, 16.0, and 17.6\%, stored as counts against the historical 256-draw field whose uniform use Appendix~\ref{sec:methods_setting} does not certify. One drawn problem was excluded because a theorem-extraction artifact affected its baseline comparison. DeepSeek-Prover's bank was selected after a cold screen of 32 bare draws on each of its 137 problems. Taking Qwen's eleven screening rates in descending order, each target received the unused DeepSeek-Prover problem with at least one observed success (in the cold screen or any one-step arm) whose cold-screen rate is closest to the target; ties go first to problems with a one-step success and then to the earliest position in a seeded random order. We reproduce these frozen model-specific banks and do not treat their outcomes as a head-to-head model comparison.

For both models, the code-bearing arms retain failed-proof history alone ($C$), with the corresponding compiler error ($CE$), with an ordinarily shuffled wrong diagnostic ($CE_s$), or with a class/length-matched wrong diagnostic ($CE_m$). The code-free arms retain corresponding-diagnostic history ($E$) or class/length-matched wrong-diagnostic history ($E_m$). Table~\ref{tab:aliases} lists the artifact label of every arm. Memoryless $I$ contains 640 instruction-only draws per problem, pooled within problem to estimate pass@$k$ using the estimator in Appendix~\ref{sec:methods_analysis}. These comparisons evaluate complete evolving-history protocols rather than one-step interventions holding a shared predecessor fixed.

\paragraph{Qwen protocol and preregistration.}
The Qwen sequential arms share a seed grid, checkpoint, chat template, decoding configuration, verifier, and software environment (vLLM 0.28.0, PyTorch 2.13, Transformers 5.16.1). Memoryless $I$, $C$, $CE$, $CE_s$, $CE_m$, and $E_m$ were generated on 6--12 September 2026 with tensor parallelism 2; $E$, noise, and assert were generated earlier, on 27--29 August with tensor parallelism 4 (Appendix~\ref{app:sequential_results}). Adaptive histories retain bounded FIFO blocks; the oldest complete blocks are evicted when the rendered history satisfies the runner's approximate threshold \texttt{characters}/3.5 $>7000$, rather than an exact tokenizer-measured cap. The history is therefore a sliding window over recent attempts. Reconstructing the rule from the logged proofs and diagnostics, a $CE$ prompt from attempt 17 onward shows a median of 7 prior attempts on Qwen (interquartile range 4--11) and 18 on DeepSeek-Prover (14--24), and $C$ shows 8 and 21; DeepSeek-Prover's blocks average about 1{,}340 characters against Qwen's 3{,}600, which is why its window holds more attempts at a similar mean prompt length (9{,}700 versus 9{,}100 tokens); code-free $E$ histories, whose blocks are short, are rarely truncated (about 2\% of Qwen and 1\% of DeepSeek-Prover prompts). Sequential diagnostic displays are capped at 1,200 characters, whereas one-step prompts show the stored diagnostic in full. A verifier crash is retried three times; one that persists is recorded as an unsuccessful attempt, and its message then enters later history like any other diagnostic (Appendix~\ref{app:integrity}). Additional Qwen controls use content-free random-word filler (noise) or a failure marker (assert). Adaptive Qwen arms begin with the bare theorem-completion prompt because no failure history exists at attempt 1, whereas memoryless $I$ includes its retry instruction from the first draw. The Qwen sensitivity analysis conditional on failure of attempt 1 addresses this initialization difference; the primary comparison does not match prompts or token expenditure.

For Qwen's $E_m$, each wrong diagnostic is drawn from another problem and matched to the corresponding own diagnostic first by coarse Lean error class and then by approximate token length. Across 4,055 wrong-diagnostic draws, 98.7\% are same-class matches; 51 use the prespecified pooled tactic-family fallback. The wrong-to-own diagnostic token-length ratio is 1.00 at both quartiles. Thus, Qwen's $E-E_m$ primarily varies problem-specific diagnostic content while approximately preserving error type and length. $CE_s$ provides an ordinarily shuffled wrong-diagnostic control in code-bearing history. Because $CE_s$ and $E_m$ use different donor policies, their combination in the registered difference-of-effects is asymmetric. The Qwen noise arm instead uses one of five fixed random-word sentences in each history block.

The $CE_m$ arm applies the same matching rule to code-bearing history, drawing from each model's completed $CE$ error pool; $E_m$ instead draws from that model's $E$ error pool. At each generated attempt after the first, the selected donor matches the preceding own error. Same-class matching succeeds for 5,662/5,755 Qwen draws and 5,002/5,793 DeepSeek-Prover draws, with 93 and 791 pooled-family fallbacks. Median longer/shorter token ratios are 1.12 and 1.13, while 90th percentiles are 1.94 and 4.43. Every retained donor comes from another problem and none is text-identical to the preceding own error. Matching is therefore approximate, and the regime-specific donor pools prevent $CE_m-E_m$ from becoming a fixed-code intervention.

The preregistered Qwen sequential primary endpoint is $CE-I$ at $k=64$. Prespecified secondary contrasts include $C-I$, $CE-C$, $CE-CE_s$, $E-E_m$, and $CE_s-E_m$. A separately preregistered Qwen difference-of-effects tests whether the estimated benefit of problem-specific diagnostic content is larger with failed-proof history than without it, subject to the differing wrong-diagnostic policies just noted:
\[
(CE-CE_s)-(E-E_m),
\]
with $k=64$ as the primary horizon. Other Qwen horizons are secondary depth-profile analyses.

The $CE_m$ registration applies separately to both models. Its primary endpoint is $CE-CE_m$ at $k=64$; registered secondary endpoints include $(CE-CE_m)-(E-E_m)$, $CE_m-I$, $CE_m-C$, and $CE_m-CE_s$. We report the ordinary-shuffle endpoints under their own labels. The unregistered matched-policy $CE_m-E_m$ comparison is exploratory.

\paragraph{Registration record.} Five timestamped registration documents cover the sequential experiments, each written before the arms it specifies were generated and each naming the generation jobs it preceded; all five are released with the artifact under \texttt{protocols/}. They cover memoryless $I$, $C$, $CE$, $CE_s$, $E_m$ and $CE_m$ on Qwen and all seven analysed arms on DeepSeek-Prover. Two Qwen arms are not covered: the code-free diagnostic history $E$ and the content-free noise control predate these registrations, so for Qwen it is the contrasts that use $E$, rather than that arm's generation, that were registered in advance. Each fixes a primary endpoint, secondary contrasts, and the analyses it treats as exploratory; Tables~\ref{tab:qwen-seq} and~\ref{tab:seq_endpoints_deepseek} label every reported contrast accordingly. DeepSeek-Prover's arms, including its ordinary-shuffle contrasts, were registered before any prover depth attempt was generated. Two departures are worth recording: one sampled Qwen problem was excluded for a theorem-extraction artifact, and the DeepSeek ordinary-shuffle intervals use 200{,}000 bootstrap resamples where the registration specified 10{,}000, which changes Monte Carlo resolution but not the estimator. Analyses introduced after the fact---the conditional-on-first-failure, leave-one-problem-out, exact-randomization and token-budget comparisons---are exploratory.

\paragraph{DeepSeek-Prover sequential experiments.}
The DeepSeek analysis uses the seven corresponding manuscript arms $I$, $C$, $CE$, $CE_m$, $CE_s$, $E$, and $E_m$ (artifact labels in Table~\ref{tab:aliases}). Its registered primary endpoint is likewise $CE-I$ at $k=64$, and it reports the same contrasts and descriptive horizons $k=1,2,4,8,16,32,64$. DeepSeek's retained memoryless prompts begin with ``Now produce a correct, complete proof.'' Its logged verifier failures remain unsuccessful draws in the pooled baseline (Appendix~\ref{app:integrity}).

\paragraph{DeepSeek-Prover sequential decoding.} Two settings differ from Qwen's. The prover ships no \texttt{generation\_config.json}, so vLLM's defaults apply and no top-$k$ or top-$p$ truncation is imposed, matching that model's one-step runs; and a tokenizer guard is asserted at job start after the chat-template defect described in Appendix~\ref{app:integrity}. The prompts also differ: as in its one-step arms, DeepSeek-Prover's theorem statement carries the Lean preamble inside its code fence (Appendix~\ref{app:prompts}). History accounting, the 1{,}200-character diagnostic cap, the eviction rule, the seed grid, and the ten-trajectory, 64-attempt protocol with early stop are as for Qwen.

\subsection{Generation and verification}
\label{sec:methods_verification}
The one-step runner requests temperature 0.7 and at most 16,384 output tokens using the model's chat template. Qwen thinking is disabled, with top-$k=20$ and top-$p=0.95$. One-step reflections use a separate generation capped at 1,024 tokens. Our primary budgets count proof attempts, not equal token expenditure.

Verification assembles the fixed preamble and dataset theorem declaration with the extracted proof body. The Kimina-based harness requests a 120-second timeout and sets a 400,000-heartbeat limit. Success requires no verification errors and no warning that the declaration uses \texttt{sorry}, including through \texttt{admit}. Unparseable outputs are unsuccessful; transient infrastructure failures are retried. Because the verifier rebuilds each proof from the dataset's theorem statement and the extracted proof body, a model cannot restate or weaken the theorem. A text scan of all 7,217 accepted proofs finds no \texttt{native\_decide}, \texttt{axiom} declaration, \texttt{implemented\_by} or \texttt{extern} attribute, \texttt{unsafe} code, macro or syntax extension, or \texttt{ofReduceBool}. The one textual \texttt{sorry} lies inside a failed \texttt{try} block, whose messages Lean discards, and that proof raises no \texttt{sorry} warning. This scan is not a per-proof \texttt{\#print axioms} audit.

\subsection{Outcomes and statistical analysis}
\label{sec:methods_analysis}
We define pass@$k$ as the probability of obtaining at least one verified proof within $k$ attempts, averaged across problems. For one-step experiments, pass@4 is estimated from independent continuations of a fixed prompt; for adaptive trajectories, we use sequential pass@$k$, the empirical fraction succeeding within their first $k$ attempts. We report paired differences in percentage points with 95\% confidence intervals and account for multiple trajectories per problem in sequential comparisons. Failed-proof copying is equality of proof bodies after removing comments and normalizing whitespace.

For sequential contrasts, inference is paired at the problem level within each model. The established Qwen endpoint intervals use 10,000 problem-bootstrap resamples. DeepSeek intervals use 200,000 resamples with NumPy \texttt{default\_rng(20260913)} (PCG64), drawing from the lexicographically sorted problem identifiers. Each replicate samples 11 problems with replacement and carries every compared arm and all trajectories or memoryless draws for each selected problem. Component effects and their difference are recomputed jointly; ten trajectories on one theorem are not treated as ten independent theorem samples. Memoryless draws are pooled within problem before resampling, without a second within-problem bootstrap. We report unadjusted percentile 95\% intervals: an interval entirely above zero supports an advantage, one entirely below zero supports underperformance, and one spanning zero resolves neither a difference nor equivalence. The inference concerns variation over the tested problem banks and is limited by having only 11 problems per model.


For repeated independent sampling, we estimate success within $k$ attempts from $c$ successes among $n$ draws as
\[
\widehat{\mathrm{pass@}k}=1-\frac{\binom{n-c}{k}}{\binom{n}{k}},
\]
then average across problems. We use $n=16$, $k=4$ for the repeated diagnostic comparison and $n=640$ for memoryless sequential controls on both models. Under identically distributed independent draws, pooling averages over all $k$-sample subsets, avoiding dependence on an arbitrary grouping into batches or trajectories.


For the pooled diagnostic comparison, we distinguish an approximate fixed-bank interval for sampling uncertainty from a paired problem-bootstrap interval that also reflects variation across problems. The fixed-bank calculation uses the pooled estimator's sampling variance and assumes independent sampling across conditions; the batch-based analysis assesses sensitivity to that assumption. Equivalence within $\pm5$ percentage points is assessed using a 90\% interval, separately from the upper bounds against instruction-only retrying.

\paragraph{Bounding feedback's benefit.}
For each tested diagnostic or reflection condition, we estimate its pass@4 difference from instruction-only retrying. We resample problems jointly across conditions within each model, preserving the paired comparisons. Besides individual intervals, we compute one-sided upper confidence limits with a Bonferroni adjustment across the 16 comparisons (nine Qwen and seven DeepSeek-Prover conditions; 100,000 bootstrap resamples). This exploratory analysis bounds the benefit compatible with the data for these tested one-step methods, rather than merely asking whether improvement is statistically detectable. Bootstrap coverage is approximate; the bounds do not apply to untested feedback methods or sequential trajectories.

\section{Prompt texts}
\label{app:prompts}
One-step prompts are assembled as \texttt{[sentence] [blocks] [instruction line] [task]}, joined by blank lines; an arm is defined by which parts are present. Sequential prompts are described separately below.

\paragraph{Task (one-step arms).}
\begin{quote}\small\ttfamily
Complete the following Lean 4 code:\\[2pt]
```lean4\\
\{statement, with the trailing sorry stripped\}\\
```
\end{quote}

\paragraph{Instruction line} (present in the one-step arms except \texttt{bare} and the \texttt{\_noI} twins): \texttt{Produce a correct, complete proof.}

\paragraph{Failure sentence} (\texttt{F} family): \texttt{You previously attempted to prove this theorem and it did not verify.}

\paragraph{Sentence variants} (post-hoc, all 13 Qwen3.6 tokens like the failure sentence). \texttt{S}: ``You previously attempted to prove this theorem and it did fully verify.'' \texttt{P}: ``Someone previously attempted to prove this theorem and it did not verify.'' \texttt{D}: ``This theorem is quite difficult and requires a careful, complete proof.'' Random-word sentences are drawn from \texttt{/usr/share/dict/words} under fixed seeds and resampled to exactly 13 tokens. \texttt{N} is ``Ennui inward evacuates catchier sequel tubing lubbers.''; the four replicate strings \texttt{N2}--\texttt{N5} and the five strings of the sequential noise arm are listed in the artifact.

\paragraph{Blocks.} Failed code: \texttt{This was the failed proof:} followed by the frozen attempt in a \texttt{lean4} fence. Error: \texttt{Lean compiler feedback from the failed attempt:} followed by the stored compiler output. In the ordinary-shuffle one-step arms, the same block contains another problem's diagnostic, sampled with replacement using seed 22072605 from bank entries whose error text differs from the corresponding diagnostic. This is not a permutation or a length-matched control; each model uses its own bank. Reflection: \texttt{Reflection on what went wrong:} followed by a model-written critique.

\paragraph{Matched one-step controls.}
The matched wrong-diagnostic arms $F_{E_m}$ and $F_{CE_m}$ use diagnostics from other problems matched by coarse error class and length. They are distinct from the ordinary-shuffle controls above; on 37/137 DeepSeek-Prover problems, a matched donor produces the same diagnostic text as the correct error (Appendix~\ref{app:ext3}). The $F_{C_o}$ control substitutes another theorem's failed proof for the model's own failed predecessor.

\paragraph{Reflection generation.} Reflections are written by the same Qwen model (temperature 0.7, seed frozen before the reruns, max 1024 tokens) from the prompt:
\begin{quote}\small\ttfamily
You are an expert Lean 4 proof coach. A proof attempt failed to verify.\\[2pt]
Theorem: \{statement\}\quad Failed proof: \{predecessor\_code\}\quad Lean compiler feedback: \{predecessor\_error\}\\[2pt]
Write a 3-5 sentence coaching critique of what went wrong and what to do differently on the next attempt. Do not write any Lean code.
\end{quote}

\paragraph{Qwen code-free history blocks.} This protocol retains one block per prior failed attempt, oldest first, with FIFO eviction under its approximate history budget (Appendix~\ref{sec:methods}). \texttt{assert}: \texttt{\#\# Previous attempt \{n\}: failed.} \texttt{noise}: the same heading followed by a random-word sentence, cycled across five strings. \texttt{f}: \texttt{\#\# Previous attempt \{n\}} followed by \texttt{Lean compiler feedback:} and that attempt's error. The replicate memoryless pool (\texttt{blind\_I}; Appendix~\ref{app:sequential_results}) carries no history and prefixes \texttt{Now produce a correct, complete proof.} to the task. The approximate history budget is defined in Appendix~\ref{sec:methods_sequential}.

\paragraph{Sequential comparisons.}
The sequential protocols are denoted $I$, $C$, $CE$, $CE_m$, $CE_s$, $E$, and $E_m$ for each model. $C$ retains failed proofs; $CE$, $CE_m$, and $CE_s$ add corresponding, class/length-matched wrong, or ordinarily shuffled wrong diagnostics; $E$ and $E_m$ retain corresponding or class/length-matched wrong diagnostics without failed proofs; $I$ carries no history. Artifact labels appear in Table~\ref{tab:aliases}. The sequential memoryless prompt prefixes the theorem-completion task with ``Now produce a correct, complete proof.'' Matching and history-budget details appear in Appendix~\ref{sec:methods_sequential}.

\paragraph{Sequential prompts for the code-bearing arms.} Attempt 1 of every sequential arm is the bare task prompt, with no failure notice, no history, and no retry instruction: \texttt{Complete the following Lean 4 code:} followed by the theorem in a \texttt{lean4} fence. Every later attempt is the concatenation of the failure notice, one block per prior failed attempt oldest first, and the retry instruction with the theorem:

\begin{quote}\small\ttfamily
You previously attempted to prove this theorem and it did not verify. Your prior attempts are summarized below.\\[2pt]
\#\# Previous attempt $n$\\
Proof attempt:\\
```lean4\\
\{that attempt's verifier input\}\\
```\\
Lean compiler feedback:\\
\{that attempt's diagnostic\}\\[2pt]
Now produce a correct, complete proof.\\
```lean4\\
\{theorem statement\}\\
```
\end{quote}

\noindent The proof block shows that attempt's verifier input: the fixed preamble, the theorem statement, and the model's proof body. $C$ omits the \texttt{Lean compiler feedback:} line and its diagnostic, keeping the proof block. $CE$ shows the attempt's own diagnostic; $CE_s$ and $CE_m$ show a wrong diagnostic in that position under their respective donor policies, leaving the rest of the block byte-identical. $E$ and $E_m$ use the code-free block, which drops the \texttt{Proof attempt:} lines and keeps only \texttt{\#\# Previous attempt $n$} and the feedback line. The retry instruction therefore appears in every attempt after the first, in every arm including $C$, so it is not a difference between the arms. Displayed diagnostics are truncated to 1{,}200 characters. Blocks are evicted oldest first while the rendered block text exceeds the runner's approximate budget (characters divided by 3.5 above 7{,}000), so the prompt grows until it saturates at that cap rather than growing without bound. For DeepSeek-Prover the Mathlib preamble is placed inside the code fence of the theorem statement, as in that model's one-step arms; Qwen's task statement carries no preamble, which is supplied to the verifier. The one-step frozen predecessors store the same verifier input, so one-step and sequential proof blocks have the same content, and in both protocols the preamble reaches a Qwen prompt only inside a displayed failed proof. Within Qwen, contrasts between code-bearing and code-free arms ($F_C-F$, $C-E$, $CE_s-E_m$, $CE_m-E_m$, and both interactions) therefore also vary whether the preamble is visible; contrasts within the code-bearing arms ($CE-C$, $CE-CE_m$, $F_{CE}-F_C$) and predecessor replacement ($F_{C_o}-F_C$, whose donor block carries the same preamble) do not. DeepSeek-Prover sees the preamble in every prompt.

\section{Statistical details}
\label{app:stats}
One-step effect sizes describe pass@4 differences in percentage points, not per-generation pass rates. Exact paired sign tests on per-problem pass counts and McNemar tests on binary success indicators are different tests; neither should be inferred from overlap of marginal figure whiskers. Appendix~\ref{sec:methods_analysis} defines the estimators and resampling units.

The preregistered six-contrast Qwen family is $F$ versus bare, $F_E$ versus $F$, $F_E$ versus bare, $F_R$ versus $F$, $F_C$ versus $F$, and $F_{CE}$ versus $F_E$. The first three sign tests are one-sided; the other three are two-sided. Holm correction applies within this family. In particular, $F_C$ versus $F$ has raw $p=0.014$ but Holm-adjusted $p=0.055$; exploratory code-content comparisons are unadjusted. The combined $F$ scaffold includes both the instruction and failure notice, not the notice alone. Two equivalence tests (TOST at $\pm8$~pp against $F$) were preregistered with this family, and neither establishes equivalence: $F_E$ versus $F$ has $p=0.14$ and $F_R$ versus $F$ has $p=0.17$, so these nulls are bounds rather than demonstrated equivalence. They are separate from the correct-versus-shuffled comparison's $\pm5$~pp margin. The pooled estimator is accompanied by the batch estimator as a sensitivity analysis (Appendix~\ref{app:precision_results}).

Main one-step recovery and interaction intervals use the diagnostic-ablation analysis (20,000 paired problem-bootstrap resamples); one-step figures show marginal intervals. The matched correspondence-by-proof-exposure and instruction-by-recovery interactions use per-problem pass@4 indicators, 20,000 paired problem-bootstrap resamples, and Python \texttt{random.Random(20260913)}. Both are exploratory and unadjusted rather than members of the preregistered family.

\paragraph{Multiplicity sensitivity.} Because several exploratory comparisons are interpreted in the text, we also adjusted them within model. Across Qwen's 15 one-step comparisons (bootstrap $p$-values; the family includes the registered $F_C-F$ and $F_{CE}-F_C$), the matched correspondence-by-proof-exposure and instruction-by-recovery interactions (each raw $p=0.011$), $F_{CE}-F_{CE_m}$ (raw $p=0.015$), $F_{CE}-F_{CE_s}$ (raw $p=0.008$), and $F_C-F$ (raw $p=0.010$) are significant after Benjamini--Hochberg adjustment (each $p_{\mathrm{BH}}\leq0.027$) but not after Holm adjustment ($0.09$ to $0.12$); $F_{CE}-F_C$, $F_{C_o}-F_C$, and the no-instruction exposure penalty survive both. Across DeepSeek-Prover's 15 one-step comparisons, the exposure penalty with and without the instruction, $F_{C_o}-F_C$, and $F_{CE}-I$ survive both adjustments, and no other comparison is significant before adjustment. For the sequential $k=64$ contrasts we use the exact randomization $p$-values of Appendix~\ref{app:sequential_results}, because percentile-bootstrap $p$-values over 11 problems are anti-conservative, and adjust over all contrasts in Tables~\ref{tab:qwen-seq} and~\ref{tab:seq_endpoints_deepseek} for each model. Qwen's $CE-CE_m$ ($p=0.004$) is the registered primary endpoint of its family and needs no family adjustment, although adjusted with the other contrasts it would survive Benjamini--Hochberg ($0.021$) but not Holm ($0.074$); $CE-C$ and $CE-CE_s$ ($p=0.012$ each) are significant after Benjamini--Hochberg ($0.035$) but not Holm ($0.19$) adjustment. On Qwen, only $CE_s-E$ and $C-E$ ($p=0.001$ and $0.002$) survive Holm adjustment. On DeepSeek-Prover, the six contrasts with $p=0.002$---$CE-I$, $C-I$, $CE_s-I$, $CE_m-I$, $CE_s-E_m$, and $CE_m-E_m$---survive Holm adjustment ($0.039$), while $E-E_m$ ($p=0.063$) and the matched interaction ($p=0.109$) are not significant before adjustment.

\paragraph{Sequential endpoint inference.}
The Qwen endpoint analysis uses 10,000 paired problem-bootstrap resamples. Its preregistered primary repair-versus-retry endpoint is $CE-I$ at $k=64$; the preregistered ordinary-shuffle interaction is $(CE-CE_s)-(E-E_m)$ at that horizon (raw artifact labels: $(CE-CE_s)-(f-E_s)$). The DeepSeek analysis uses 200,000 paired problem-bootstrap resamples, NumPy \texttt{default\_rng(20260913)} (PCG64), and lexicographically sorted problem identifiers. In each model, a replicate samples 11 problems with replacement and carries all compared arms and all trajectories or pooled memoryless values jointly. Component effects and their difference are recomputed within the same replicate. Bootstrap sampling is within model, not across models or independently across arms.

The separately registered $CE_m$ family is analyzed for each model with 10,000 paired problem-bootstrap resamples using Python \texttt{random.Random(0)} and the retained bank order. Its primary contrast is $CE-CE_m$; the matched interaction $(CE-CE_m)-(E-E_m)$ and $CE_m-I$, $CE_m-C$, and $CE_m-CE_s$ are registered secondary contrasts. The unregistered $CE_m-E_m$ comparison is exploratory. These intervals remain separate from the analyses above, whose resampling procedures and ordinary-shuffle endpoints are also reported.

The sequential intervals are unadjusted 95\% percentile intervals. Intervals entirely above or below zero resolve an advantage or underperformance, respectively; intervals spanning zero resolve neither a difference nor equivalence. Other horizons are descriptive or secondary; the Qwen repetition and conditional-on-attempt-1-failure analyses remain Qwen-specific. Neither the sign of an unresolved interaction nor an interval excluding zero at an intermediate horizon establishes a positive interaction at $k=64$.

\paragraph{Sequential figure bands.}
Figure~\ref{fig:sequential} uses the Qwen plotting procedure (2,000 problem resamples, Python \texttt{random.Random(0)}) and uses the 200,000-resample DeepSeek procedure above. Left-panel bands are marginal intervals for each arm; right-panel bands are paired intervals for the two diagnostic contrasts and their difference. All are pointwise 95\% intervals, not simultaneous confidence bands over depth. The Qwen plotting bands and separately reported 10,000-resample endpoint intervals need not have identical bounds. Inferences use the paired endpoint contrasts in the text, not overlap of marginal curves.

Memoryless $I$ is the per-problem pooled pass@$k$ estimate from 640 draws for both models, not the pass fraction of arbitrarily assigned trajectory groups. Adaptive curves use the observed fraction of ten trajectories per problem with a first pass by $k$. Resampling does not separately estimate all rollout uncertainty conditional on a fixed problem bank. The two model banks differ, so resolved recovery on Qwen and an unresolved recovery contrast on DeepSeek do not themselves establish a causal difference between models.
\paragraph{Claims and their evidential status.} Table~\ref{tab:claims} lists the contrasts that carry the main-text claims, with their registration status and how they fare under the multiplicity adjustments above.

\begin{table}[h]
\centering
\caption{Main-text claims, the contrasts that carry them, and their status. Intervals are the paper's 95\% paired problem-bootstrap intervals. Status: Prim.\ = registered primary endpoint; Reg.\ = registered secondary or family member; Expl.\ = exploratory. The last column gives the adjustments a contrast survives, applying Holm and Benjamini--Hochberg (BH) within model as described above (one-step: bootstrap $p$ over each model's 15 comparisons; sequential: exact randomization $p$ over each model's table); ``---'' marks contrasts for which no difference is claimed; primary endpoints are shown as if adjusted, although they need no adjustment. Qwen's preregistered $F_C-F$ has Holm-adjusted $p=0.055$; DeepSeek-Prover's $E-E_m$ and matched interaction have exact $p=0.063$ and $0.109$ before adjustment. DS = DeepSeek-Prover.}
\label{tab:claims}
\small
\begin{tabular}{llrlll}
\toprule
Model & Contrast & $\Delta$ (pp) & 95\% CI & Status & Adjustment \\
\midrule
\multicolumn{6}{l}{\emph{Repair does not beat retrying}} \\
Qwen & $F_{CE}-I$ & $-3.0$ & $[-8.4,+2.4]$ & Expl. & --- \\
DS & $F_{CE}-I$ & $-16.1$ & $[-22.6,-10.2]$ & Expl. & Holm, BH \\
Qwen & $CE-I$ & $-18.3$ & $[-41.8,+5.3]$ & Prim. & --- \\
DS & $CE-I$ & $-50.7$ & $[-66.4,-34.5]$ & Prim. & Holm, BH \\
\midrule
\multicolumn{6}{l}{\emph{Own failed proof harms}} \\
Qwen & $F_C-F$ & $-9.0$ & $[-15.6,-2.4]$ & Reg. & BH only \\
DS & $F_C-F$ & $-15.3$ & $[-21.9,-9.5]$ & Expl. & Holm, BH \\
Qwen & $F_{C_o}-F_C$ & $+14.4$ & $[+7.8,+21.6]$ & Expl. & Holm, BH \\
DS & $F_{C_o}-F_C$ & $+13.9$ & $[+8.0,+20.4]$ & Expl. & Holm, BH \\
Qwen & $C-I$ & $-40.1$ & $[-62.2,-17.0]$ & Reg. & BH only \\
DS & $C-I$ & $-50.7$ & $[-67.6,-33.6]$ & Reg. & Holm, BH \\
\midrule
\multicolumn{6}{l}{\emph{Correct diagnostic helps within repair (Qwen)}} \\
Qwen & $F_{CE}-F_C$ & $+13.2$ & $[+7.2,+19.2]$ & Expl. & Holm, BH \\
Qwen & $F_{CE}-F_{CE_m}$ & $+7.8$ & $[+1.8,+13.8]$ & Expl. & BH only \\
Qwen & $CE-CE_m$ & $+23.6$ & $[+12.7,+35.5]$ & Prim. & BH only \\
Qwen & $CE-C$ & $+21.8$ & $[+10.9,+32.7]$ & Reg. & BH only \\
Qwen & $CE-CE_s$ & $+18.2$ & $[+8.2,+28.2]$ & Reg. & BH only \\
DS & $CE-CE_m$ & $+1.8$ & $[-5.5,+9.1]$ & Prim. & --- \\
\midrule
\multicolumn{6}{l}{\emph{No correctness advantage without the failed proof}} \\
Qwen & $F_E-F_{E_s}$, pooled & $+1.73$ & $[-2.06,+5.62]$ & Reg. & --- \\
Qwen & $E-E_m$ & $+6.4$ & $[-6.4,+19.1]$ & Reg. & --- \\
DS & $E-E_m$ & $-14.5$ & $[-30.0,-2.7]$ & Reg. & none \\
\midrule
\multicolumn{6}{l}{\emph{Interactions unresolved}} \\
Qwen & $(CE-CE_m)-(E-E_m)$ & $+17.3$ & $[-1.8,+34.5]$ & Reg. & --- \\
DS & $(CE-CE_m)-(E-E_m)$ & $+16.4$ & $[+0.9,+35.5]$ & Reg. & none \\
Qwen & one-step, matched & $+11.38$ & $[+2.40,+19.76]$ & Expl. & BH only \\
Qwen & one-step, instruction & $+10.18$ & $[+2.40,+17.96]$ & Expl. & BH only \\
\bottomrule
\end{tabular}
\end{table}

\section{Detailed results and analyses}
\label{app:results_details}
\label{app:tables}

\subsection{One-step results and framing ablations}
\label{app:framing_results}
\begin{figure}[t]
\centering
\includegraphics[width=\linewidth]{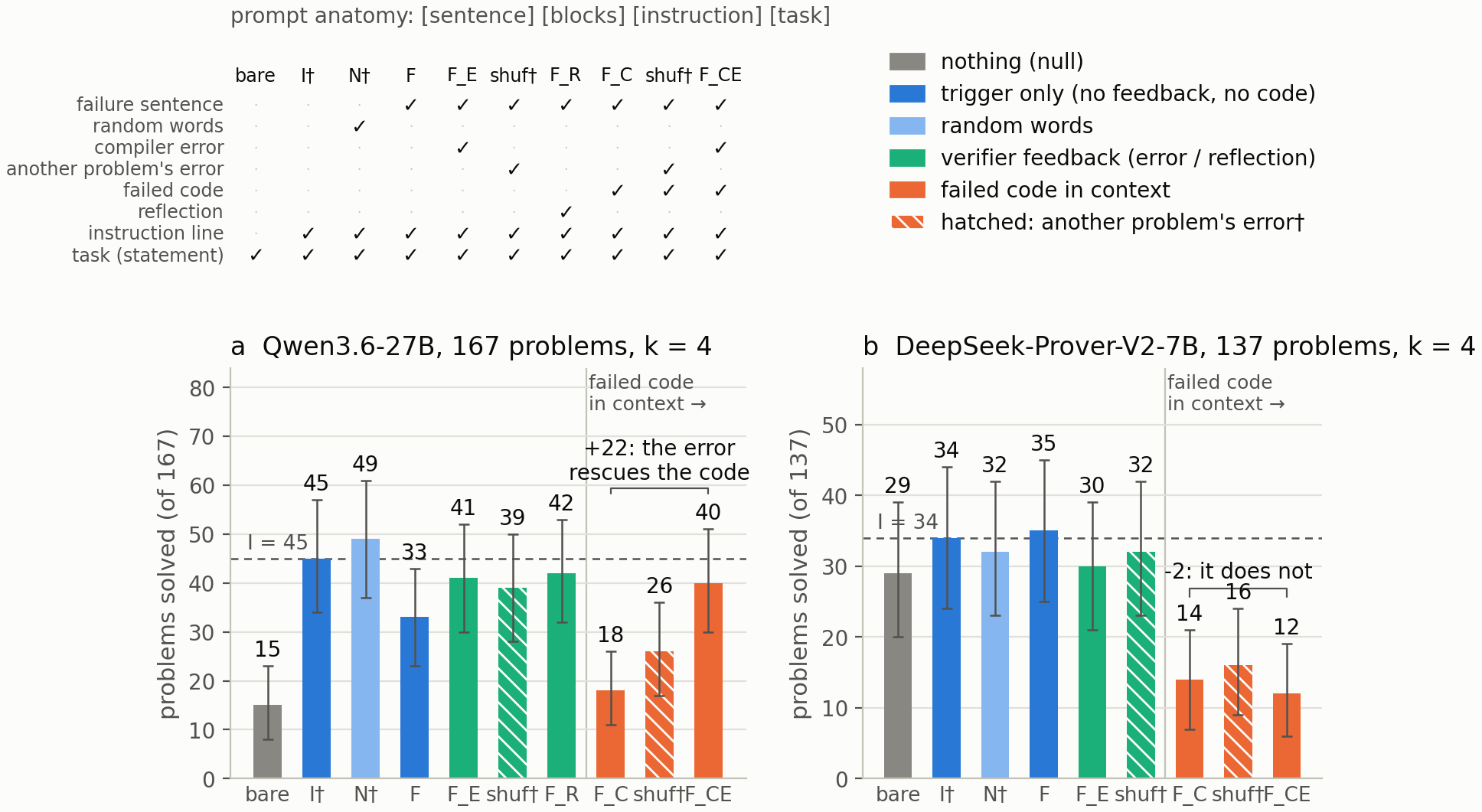}
\caption{One-step pass@4 for the central arms, and prompt anatomy. Dashed lines mark instruction-only performance; hatching denotes shuffled wrong diagnostics, and the bracket gives the $F_{CE}-F_C$ difference in problems solved. Whiskers are marginal 95\% problem-bootstrap intervals, not paired contrasts. The predecessor-replacement and matched-diagnostic controls are summarized in Fig.~\ref{fig:decomposition} and Table~\ref{tab:ext3}; detailed copying and diversity appear in Fig.~\ref{fig:copy_details}.}
\label{fig:decomposition_details}
\end{figure}

Figure~\ref{fig:decomposition_details} shows the full one-step arm comparison and prompt anatomy.

Similar aggregate counts conceal different coverage: instruction-only and failure-notice-only retrying share 38 successful problems, with a union of 53, despite individually reaching 45 and 46. This is a descriptive comparison using both sets of samples, not an equal-budget improvement.

The failure notice alone and instruction alone yield similar aggregate Qwen outcomes (46 and 45 of 167 problems), but combining them yields 33. A fresh-seed failure-notice replicate reaches 45; its comparison against the combined scaffold remains favorable (paired pass-count sign $p=0.017$). Thus, the lower performance of $F$ is not evidence that the failure notice alone is a weak retry prompt.

\begin{figure}[t]
\centering
\includegraphics[width=\linewidth]{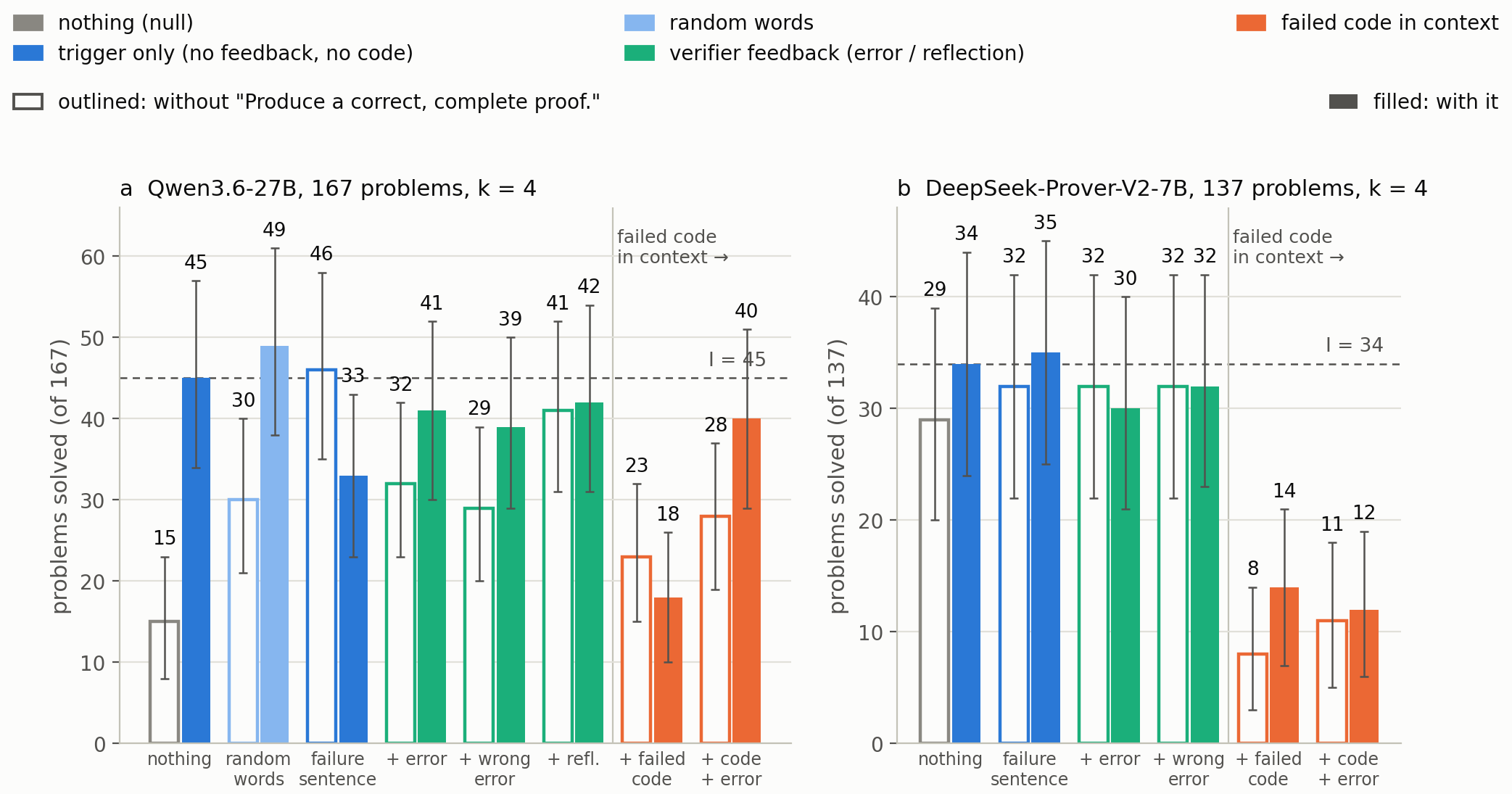}
\caption{Instruction ablations. Outlined/filled bars omit/include the instruction line; $F$ without that line is the failure notice alone. Bars show pass@4 counts with marginal problem-bootstrap intervals.}
\label{fig:framing_details}
\end{figure}

The diagnostic's apparent contribution depends on this surrounding text (Fig.~\ref{fig:framing_details}). Adding the correct diagnostic to $F$ raises the count from 33 to 41; adding it to the failure notice without the instruction lowers it from 46 to 32. A shuffled diagnostic without the instruction reaches 29. DeepSeek-Prover reaches 32 for the notice alone, correct diagnostic, and shuffled diagnostic in the no-instruction setting. These are distinct comparisons, not one diagnostic effect relative to an interchangeable retry baseline.

Qwen random-word-plus-instruction variants reach 49, 48, 43, 42, and 44 problems; these observations establish sensitivity to framing, not that arbitrary random text reliably improves performance. Reflection reaches 42 with the instruction and 41 without it. Removing the instruction does not eliminate the failed-proof penalty: Qwen reaches 23 with code alone and 28 with code plus diagnostics, versus 46 for the notice alone; DeepSeek reaches 8 and 11 versus 32. Full arm counts appear in Tables~\ref{tab:qwen}--\ref{tab:prover}.
\begin{table}[h]\centering\small
\caption{Qwen3.6-27B, all one-step arms (167 problems; 4 samples per problem). ``Problems passed'' = problems with $\geq 1$ passing sample (pass@4 count); ``passes'' = passing samples of 668. $^\dagger$ post-hoc.}
\label{tab:qwen}
\begin{tabular}{llrrl}
\hline
arm & prompt & pass@4 count & passes & vs.\ $I$ (sign, $p$) \\ \hline
bare & task only & 15 & 18 & 9/42, $3.4\times10^{-6}$ \\
$I^\dagger$ & instruction line + task & 45 & 110 & --- \\
N$^\dagger$ & random words + instruction & 49 & 105 & 19/23, 0.64 \\
N2--N5$^\dagger$ & four more random-word sentences & 48/43/42/44 & 130/87/117/116 & 0.16/0.20/1/0.50 \\
$F$ & failure sentence + instruction & 33 & 81 & 11/26, 0.020 \\
S$^\dagger$ & ``did fully verify'' variant & 30 & 77 & 8/27, 0.0019 \\
P$^\dagger$ & ``Someone'' variant & 30 & 56 & 11/30, 0.0043 \\
D$^\dagger$ & difficulty cue, no failure claim & 35 & 81 & 11/27, 0.014 \\
$F_E$ & + compiler error & 41 & 89 & 17/28, 0.14 \\
$F_{E_s}^\dagger$ & + another problem's error & 39 & 82 & 13/25, 0.073 \\
$F_R$ & + reflection & 42 & 118 & 25/19, 0.45 \\
$F_C$ & + failed code & 18 & 36 & 6/39, $5\times10^{-7}$ \\
$F_{CE}$ & + failed code + error & 40 & 106 & 21/23, 0.88 \\
$F_{CE_s}^\dagger$ & + failed code + wrong error & 26 & 54 & 13/36, 0.0014 \\ \hline
\end{tabular}
\end{table}

\begin{table}[h]\centering\small
\caption{Qwen3.6-27B, the no-instruction ladder (167 problems). Each scaffold with and without the instruction line; prompts differ only by the removed line, on frozen banks and seeds.}
\label{tab:noi}
\begin{tabular}{lccll}
\hline
scaffold & without line & with line & line's worth (sign, $p$) & without, vs.\ bare \\ \hline
nothing & 15 (bare) & 45 ($I$) & $+30$ (42/9, $3.4\times10^{-6}$) & --- \\
random words & 30 & 49 & $+19$ (33/12, 0.002) & $+15$ (26/8, 0.003) \\
failure sentence & 46 (repl.\ 45) & 33 & $-13$ (14/29, 0.03) & $+31$ (42/7, $4\times10^{-7}$) \\
+ compiler error & 32 & 41 & $+9$ (22/11, 0.08; McNemar 0.049) & $+17$ (28/8, 0.001) \\
+ wrong error & 29 & 39 & $+10$ (25/13, 0.07) & $+14$ (28/11, 0.009) \\
+ reflection & 41 & 42 & $+1$ (18/18, 1) & $+26$ (39/10, $4\times10^{-5}$) \\
+ failed code & 23 & 18 & $-5$ (10/15, 0.42) & $+8$ (22/11, 0.08) \\
+ code + error & 28 & 40 & $+12$ (28/13, 0.03) & $+13$ (25/9, 0.009) \\ \hline
\end{tabular}
\end{table}

\begin{table}[h]\centering\small
\caption{DeepSeek-Prover-V2-7B, all arms (137 problems, its own frozen failures; four samples per problem). Entries are problems with a passing sample (pass@4 count), with passing samples out of 548 in parentheses.}
\label{tab:prover}
\begin{tabular}{lcc}
\hline
scaffold & with line & without line \\ \hline
nothing & 29 (65; bare) $\cdot$ 34 (85; $I$) & --- \\
random words & 32 (83) & --- \\
failure sentence & 35 (92) & 32 (88) \\
+ compiler error & 30 (79) & 32 (81) \\
+ another problem's error & 32 (83) & 32 (86) \\
+ failed code & 14 (25) & 8 (12) \\
+ failed code + error & 12 (27) & 11 (22) \\
+ failed code + wrong error & 16 (29) & --- \\ \hline
\end{tabular}
\end{table}

\subsection{Diagnostic precision and feedback-benefit bounds}
\label{app:precision_results}
Table~\ref{tab:primary_contrasts} reports the preregistered six-contrast family separately from the diagnostic-content and instruction-only comparisons.

\begin{table}[h]
\centering\small
\caption{Preregistered Qwen comparison family. Counts are problems with a pass among four; wins/losses compare per-problem numbers of passing samples. The first three tests are one-sided and the last three two-sided; Holm adjustment covers all six.}
\label{tab:primary_contrasts}
\begin{tabular}{lrrrr}
\hline
Contrast & Counts & Wins/losses & Raw $p$ & Holm $p$\\ \hline
$F$ vs.\ bare & 33/15 & 32/9 & 0.000215 & 0.00108\\
$F_E$ vs.\ $F$ & 41/33 & 25/19 & 0.226 & 0.462\\
$F_E$ vs.\ bare & 41/15 & 37/9 & 0.0000203 & 0.000122\\
$F_R$ vs.\ $F$ & 42/33 & 25/15 & 0.154 & 0.462\\
$F_C$ vs.\ $F$ & 18/33 & 11/27 & 0.0139 & 0.0554\\
$F_{CE}$ vs.\ $F_E$ & 40/41 & 25/22 & 0.771 & 0.771\\ \hline
\end{tabular}
\end{table}

Pooling the 16 additional code-free Qwen draws yields pass@4 estimates of 22.53\% for correct diagnostics and 20.79\% for shuffled diagnostics, a difference of $+1.73$ pp. The paired problem-bootstrap 95\% interval is $[-2.06,+5.62]$, so the experiment does not resolve an advantage across problems. Conditional on the fixed 167-problem bank, approximate 95\% and 90\% intervals are $[-0.58,+4.05]$ and $[-0.21,+3.68]$; the latter lies inside the $\pm5$ pp practical-equivalence margin. The fixed-bank calculation assumes independent, identically distributed draws within each condition and independent sampling across conditions, so we treat it as secondary to problem-level uncertainty.

The four-batch estimator gives $+1.65$ pp, fixed-bank 95\% $[-0.75,+4.04]$, and 90\% $[-0.36,+3.66]$. Its block-level sensitivity interval is wider, $[-3.54,+6.83]$, reflecting a different treatment of batch variation. Two unresolved infrastructure rows remain unsuccessful, matching the source analysis.

The benefit bound against instruction-only retrying is separate. Its family comprises nine Qwen conditions ($F_E$, $F_{E_s}$, $F_{CE}$, $F_{CE_s}$, the three available no-instruction diagnostic variants, and two reflection variants) and seven DeepSeek conditions (the same diagnostic variants, without reflection). The no-instruction variants are diagnostic-only correct, diagnostic-only shuffled, and code plus correct diagnostic. Each is compared with its model's $I$ baseline. Joint problem resampling preserves pairing; the one-sided percentile is adjusted by $0.05/16$. Individual two-sided 95\% upper endpoints are no greater than $+3.59$ pp, and family-adjusted one-sided upper limits are no greater than $+5.99$ pp. These approximate exploratory limits are neither exact finite-sample guarantees nor posterior probability statements.

\paragraph{Per-sample success.} Pass@4 counts a problem once however many of its four samples pass, so we also compared per-sample pass rates with instruction-only retrying, paired by problem (10{,}000 problem-bootstrap resamples, exploratory and unadjusted). No diagnostic or reflection condition shows a resolved per-sample advantage over $I$ on either model. On Qwen the largest are reflection ($+1.2$~pp; $[-1.8,+4.2]$), reflection without the instruction ($+0.4$~pp; $[-2.8,+3.6]$), and code plus correct diagnostic ($-0.6$~pp; $[-4.6,+3.4]$). On DeepSeek-Prover the largest among diagnostic conditions is the matched wrong diagnostic ($+0.4$~pp; $[-2.2,+2.9]$), and every condition showing the model's own failed proof trails by 10 to 13~pp (13.3~pp for code alone without the instruction). Among content-free prompts, one of the five random-word-plus-instruction strings exceeds $I$ on Qwen ($+3.0$~pp; $[+0.1,+6.0]$), which we read as string-level sensitivity rather than a benefit of arbitrary text (Appendix~\ref{app:withdrawn}).

\paragraph{Problems the verifier cannot parse.}
\label{app:parseable}
Thirty one-step problems, in both banks, have dataset statements that our verifier's Mathlib cannot parse (Appendix~\ref{sec:methods_setting}); no condition solves them, so they add zero to every paired contrast and dilute every effect by the share of the bank they occupy. Table~\ref{tab:parseable} repeats the main one-step contrasts on the 137 Qwen and 107 DeepSeek-Prover problems that parse (10{,}000 problem-bootstrap resamples). Effects grow by roughly a fifth and no conclusion changes. The pooled correct-minus-shuffled precision estimate becomes $+2.11$~pp (problem bootstrap $[-2.55,+6.84]$), with a fixed-bank 90\% interval of $[-0.26,+4.48]$, still inside the $\pm5$~pp margin; the benefit bounds against instruction-only retrying become $+4.38$~pp (largest individual upper limit) and $+7.30$~pp (family-adjusted).

\begin{table}[h]
\centering
\caption{One-step contrasts (pass@4, pp) on all problems and on the problems whose statements parse. Both columns use the same paired problem bootstrap (10{,}000 resamples), so an all-problem interval can differ by one resampling step from the main-text value for the same contrast.}
\label{tab:parseable}
\small
\begin{tabular}{llll}
\toprule
Model & Contrast & All problems & Parseable problems \\
\midrule
Qwen & $F_{CE}-I$ & $-3.0$ $[-8.4,+2.4]$ & $-3.6$ $[-10.2,+2.9]$ \\
 & $F_C-F$ & $-9.0$ $[-15.6,-2.4]$ & $-10.9$ $[-19.0,-2.9]$ \\
 & $F_{CE}-F_C$ & $+13.2$ $[+7.2,+19.8]$ & $+16.1$ $[+8.8,+23.4]$ \\
 & $F_{CE}-F_{CE_m}$ & $+7.8$ $[+1.8,+13.8]$ & $+9.5$ $[+2.2,+16.8]$ \\
 & $F_E-F_{E_m}$ & $-3.6$ $[-9.0,+1.8]$ & $-4.4$ $[-10.9,+2.2]$ \\
 & $F_{C_o}-F_C$ & $+14.4$ $[+7.8,+21.6]$ & $+17.5$ $[+9.5,+25.5]$ \\
 & matched interaction & $+11.4$ $[+3.0,+20.4]$ & $+13.9$ $[+3.6,+24.8]$ \\
 & instruction interaction & $+10.2$ $[+2.4,+18.0]$ & $+12.4$ $[+2.9,+21.9]$ \\
DeepSeek & $F_{CE}-I$ & $-16.1$ $[-22.6,-10.2]$ & $-20.6$ $[-28.0,-13.1]$ \\
 & $F_C-F$ & $-15.3$ $[-21.9,-9.5]$ & $-19.6$ $[-27.1,-12.1]$ \\
 & $F_{CE}-F_C$ & $-1.5$ $[-5.1,+2.2]$ & $-1.9$ $[-6.5,+2.8]$ \\
 & $F_{CE}-F_{CE_m}$ & $0.0$ $[-3.6,+3.6]$ & $0.0$ $[-4.7,+4.7]$ \\
 & $F_{C_o}-F_C$ & $+13.9$ $[+8.0,+20.4]$ & $+17.8$ $[+10.3,+25.2]$ \\
\bottomrule
\end{tabular}
\end{table}

\subsection{Predecessor-replacement and matched-diagnostic controls}
\label{app:ext3}

Table~\ref{tab:ext3} reports the replacement and matched-diagnostic controls using the same frozen problem banks and four samples per problem as the corresponding one-step arms.

\begin{table}[h]\centering\small
\caption{Replacement and matched-diagnostic one-step controls. ``Problems'' is the number solved at least once among four samples; ``passes'' is the number of verified passing samples. $C_o$ replaces the model's own failed predecessor with another theorem's failed proof; $E_m$ uses a wrong diagnostic matched by coarse error class and length.}
\label{tab:ext3}
\begin{tabular}{lrr|rr}
\hline
& \multicolumn{2}{c|}{Qwen (167)} & \multicolumn{2}{c}{DeepSeek-Prover (137)}\\
arm & problems & passes & problems & passes\\ \hline
$F_{C_o}$ & 42 & 93 & 33 & 88\\
$F_{E_m}$ & 47 & 90 & 32 & 87\\
$F_{CE_m}$ & 27 & 66 & 12 & 26\\ \hline
\end{tabular}
\end{table}

Replacing the own predecessor with another theorem's failed proof improves success relative to $F_C$ by $+14.4$ pp on Qwen (42 versus 18; 95\% paired-bootstrap interval $[+7.8,+21.6]$) and $+13.9$ pp on DeepSeek-Prover (33 versus 14; $[+8.0,+20.4]$). Relative to the no-code scaffold $F$, the unrelated-proof arm is unresolved on both models (Qwen: $+5.4$ pp, $[-1.2,+12.0]$; DeepSeek-Prover: $-1.5$ pp, $[-5.8,+2.2]$). This supports predecessor-specific harm in the tested comparison rather than a generic proof-context explanation.

Matching wrong diagnostics by error class and length does not reveal a code-free correctness advantage. On Qwen, $F_{E_m}$ solves 47 problems versus 41 for $F_E$ ($+3.6$ pp; $[-1.8,+9.0]$) and 39 for $F_{E_s}$ ($+4.8$ pp; $[-0.6,+10.2]$); DeepSeek-Prover solves 32, 30, and 32. With Qwen's own failed proof, $F_{CE_m}$ solves 27 problems versus 40 for $F_{CE}$: correct-minus-matched is $+7.8$ pp ($[+1.8,+13.8]$). Matched and shuffled wrong diagnostics show no resolved difference ($+0.6$ pp; $[-4.8,+6.0]$). For DeepSeek-Prover, correct, matched, and shuffled code-bearing diagnostics solve 12, 12, and 16 problems. On 37/137 prover problems the matched wrong diagnostic is byte-identical to the correct diagnostic because common syntax errors share the same message; this weakens the matched-error intervention for that model but does not affect the independently shuffled control.

\paragraph{Excluding the byte-identical matched donors.} Because common Lean syntax errors share a message, the class- and length-matching procedure returns a donor whose text equals the correct diagnostic on 37 of 137 DeepSeek-Prover problems and on 1 of 167 Qwen problems. The matcher is a one-to-one minimum-cost assignment within each error class, so each bank diagnostic serves as a donor exactly once; identical text carries a large penalty but cannot be avoided when most of a class shares one message. Twenty-nine of the pairs are statement-parse errors (\texttt{unexpected token 'in'; expected ','}) on problems whose dataset statement the verifier cannot parse (Appendix~\ref{sec:methods_setting}), so these diagnostics describe the statement rather than the model's proof; 5 more come from 8 of 11 no-goals predecessors sharing one message, 1 from the resource class, and 2 from two predecessors that store the same verifier-crash message. On those problems $F_{CE}$ and $F_{CE_m}$ are byte-identical prompts, so any observed difference is sampling noise and the intervention is null by construction. Excluding them leaves DeepSeek-Prover's matched contrast at exactly $0.0$~pp ($[-5.0,+5.0]$, versus $0.0$~pp $[-3.6,+3.6]$ on all 137); $F_E-F_{E_m}$ moves from $-1.5$~pp ($[-6.6,+3.6]$) to $-3.0$~pp ($[-10.0,+3.0]$); and the exploratory matched interaction moves from $+1.5$~pp ($[-5.1,+8.0]$) to $+3.0$~pp ($[-6.0,+12.0]$). All three remain unresolved. The 37 excluded problems are also the hardest in the bank: $F_{CE}$ and $F_{CE_m}$ each solve 1 of 37, $F$ solves 2, and 34 of the 37 (including the 29 unparseable ones) are solved under no one-step condition, so they carry almost no signal in either direction, which is why removing them widens the interval while leaving the point estimate at zero. On the 37, the correct-minus-matched difference is exactly zero, as the identical prompts require.

\paragraph{How the replacement predecessor was chosen.} $F_{C_o}$ reuses the seeded cross-problem partner that the ordinary-shuffle diagnostic arms draw from, so its donor is random with respect to the properties of the model's own predecessor. The donor shares the own predecessor's coarse error class on 40 of 167 Qwen problems and 6 of 137 DeepSeek-Prover problems. The replacement-to-own character-length ratio has quartiles $0.49/1.12/2.30$ on Qwen (range $0.02$--$21.73$) and $0.62/1.06/1.76$ on DeepSeek-Prover (range $0.11$--$6.79$); median predecessor lengths are 640 characters (own) versus 778 (replacement) on Qwen and 821 versus 877 on DeepSeek-Prover. Six Qwen donors are drawn from the four framing-holdout problems excluded from the analysed bank, since the donor pool is the full 171-problem bank. The replacement benefit is nevertheless similar on the better-matched subsets: restricting to same-class donors gives $+20.0$~pp ($[+5.0,+35.0]$, $n=40$) on Qwen and $+16.7$~pp ($[0.0,+50.0]$, $n=6$) on DeepSeek-Prover, and restricting to donors within a $0.8$--$1.25$ length ratio gives $+13.0$~pp ($[-8.7,+34.8]$, $n=23$) and $+11.4$~pp ($[+2.9,+22.9]$, $n=35$), against $+14.4$~pp and $+13.9$~pp on the full banks. The subsets are small and two of the four intervals span zero, so this is a consistency check rather than a matched experiment; a donor pool matched on length and error class is the cleaner test and is listed as future work.

\paragraph{Diagnostic match quality and its sensitivity.} Matching is a deterministic minimum-cost assignment within error class on $|\log$ token ratio$|$, with identical text penalized, and it leaves other properties of the message uncontrolled; we therefore measured those per problem and re-estimated the contrasts on the best-matched pairs. Every donor is same-class and cross-problem on both banks. The token-length ratio between the own and donor diagnostics, taken larger-over-smaller so it is symmetric and at least one, has median 1.26 on Qwen with a 90th percentile of 5.57, and median 1.05 on DeepSeek-Prover with a 90th percentile of 2.21. Lexical overlap between the own and donor message, measured as Jaccard overlap of identifier tokens, has median 0.143 on Qwen and 0.308 on DeepSeek-Prover, reflecting shared Lean boilerplate rather than shared problem content; on DeepSeek-Prover its upper quartile reaches 1.000 because of the byte-identical donors. Restricting Qwen's code-bearing contrast to better-matched pairs keeps the point estimate in the same range but at the edge of resolution: against non-identical donors within a 2.0, 1.5, and 1.25 token-length ratio it is $+7.3$~pp ($[0.0,+14.5]$, $n=110$), $+7.7$~pp ($[-1.1,+16.5]$, $n=91$), and $+9.9$~pp ($[+1.2,+19.8]$, $n=81$), and against donors whose lexical overlap with the correct message is below the bank median it is $+12.3$~pp ($[+6.2,+19.8]$, $n=81$), compared with $+7.8$~pp on the full bank. The effect is therefore not produced by the loosely matched donors, but its precision against the closest-matched donors is limited. On the 110 pairs within a 2.0 ratio, the exploratory matched interaction is $+14.5$~pp ($[+3.6,+25.5]$), about half of it from the code-free contrast $F_E-F_{E_m}=-7.3$~pp ($[-14.5,0.0]$). On DeepSeek-Prover the contrast remains unresolved near zero on every subset. Matching still does not control source locations, sub-error counts, or the diagnostic's specificity, so ``correspondence'' here means lexical and structural relevance to the failed proof rather than demonstrated semantic understanding.

\paragraph{Both models on the shared problems.} The two one-step banks are not independent: DeepSeek-Prover's 137 problems are a subset of Qwen's 167, and Qwen ran every condition on all of them, so the models can be compared on a common problem set. Doing so removes bank-selection confounding but not predecessor confounding, since each model still conditions on its own failed predecessor for a shared problem; the comparison is exploratory. On the shared 137, both results the paper treats as common to the two models remain resolved. Exposing the own failed proof costs $-7.3$~pp on Qwen ($[-13.1,-1.5]$) and $-15.3$~pp on DeepSeek-Prover ($[-21.9,-9.5]$) relative to the failure-notice scaffold, and replacing that predecessor with another theorem's failed proof recovers $+11.7$~pp ($[+5.8,+18.2]$) and $+13.9$~pp ($[+8.0,+20.4]$). The diagnostic recovery remains Qwen-only: $F_{CE}-F_C$ is $+9.5$~pp ($[+3.6,+15.3]$) on Qwen and $-1.5$~pp ($[-5.1,+2.2]$) on DeepSeek-Prover. Qwen's correct-versus-matched contrast is $+5.1$~pp on the shared subset ($[+0.0,+10.9]$) against $+7.8$~pp on the full bank, so on 30 fewer problems it no longer excludes zero; we report the full-bank estimate as the primary one and this as a precision-limited subset check. The repair-versus-retry comparison also holds on the shared problems: $F_{CE}-I$ is $-1.5$~pp ($[-7.3,+4.4]$) on Qwen and $-16.1$~pp ($[-22.6,-10.2]$) on DeepSeek-Prover, so neither model's complete repair condition beats instruction-only resampling there either. The exposure penalty does not, however, land on the same problems in the two models: of the 30 problems where it costs at least one model a solve, only 5 cost both.

Two added Qwen interactions are exploratory and unadjusted. The matched diagnostic-correspondence interaction, $(F_{CE}-F_{CE_m})-(F_E-F_{E_m})$, is $+11.38$ pp with a 95\% paired problem-bootstrap interval of $[+2.40,+19.76]$. The retry-instruction interaction, $(F_{CE}-F_C)-(F_{CE}^{\mathrm{noI}}-F_C^{\mathrm{noI}})$, is $+10.18$ pp ($[+2.40,+17.96]$). These quantify component dependence in the retained one-step bank; they were not preregistered sequential endpoints.

\subsection{Copying, diversity, and diagnostic-responsive behavior}
\label{app:behavior_results}
\begin{figure}[t]
\centering
\includegraphics[width=\linewidth]{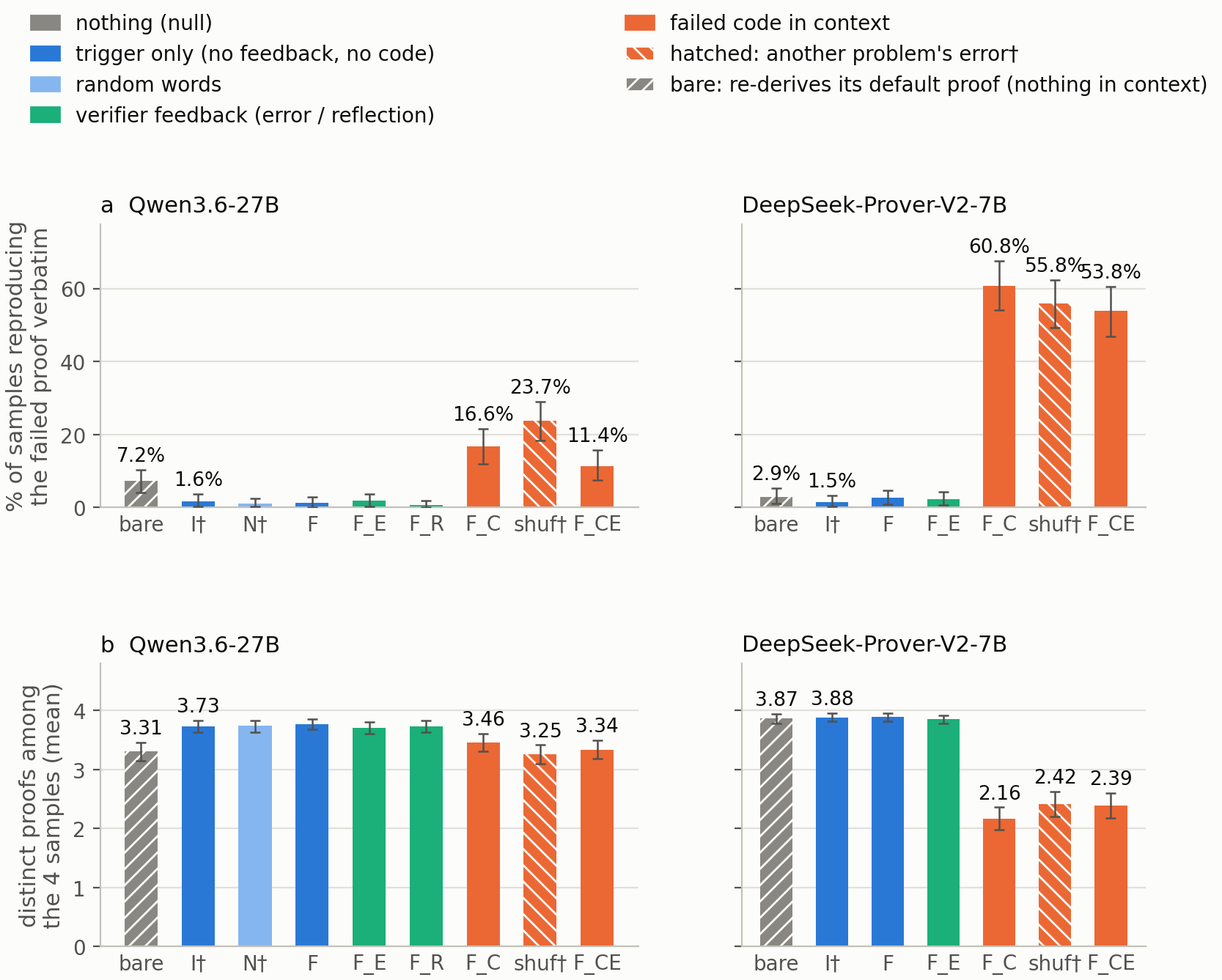}
\caption{Detailed copying and diversity across one-step conditions. Top: fraction of samples matching the normalized predecessor. Bottom: mean distinct normalized proof bodies among four samples. Hatching marks shuffled wrong diagnostics, except on the gray bare bars, whose prompt shows no predecessor. DeepSeek's bare copying rate excludes its selected predecessor draw, so it is not directly comparable with the four-draw instruction-only rate. Whiskers are marginal problem-bootstrap intervals.}
\label{fig:copy_details}
\end{figure}

Figure~\ref{fig:copy_details} expands the copying comparison in Fig.~\ref{fig:copy} and reports within-condition proof diversity.

The seven content-free retry prompts ($I$, the five random-word-plus-instruction variants, and the failure notice alone) collectively cover 69/167 problems. This uses all seven sets of samples and is not a same-budget baseline.

\begin{figure}[t]
\centering
\includegraphics[width=\linewidth]{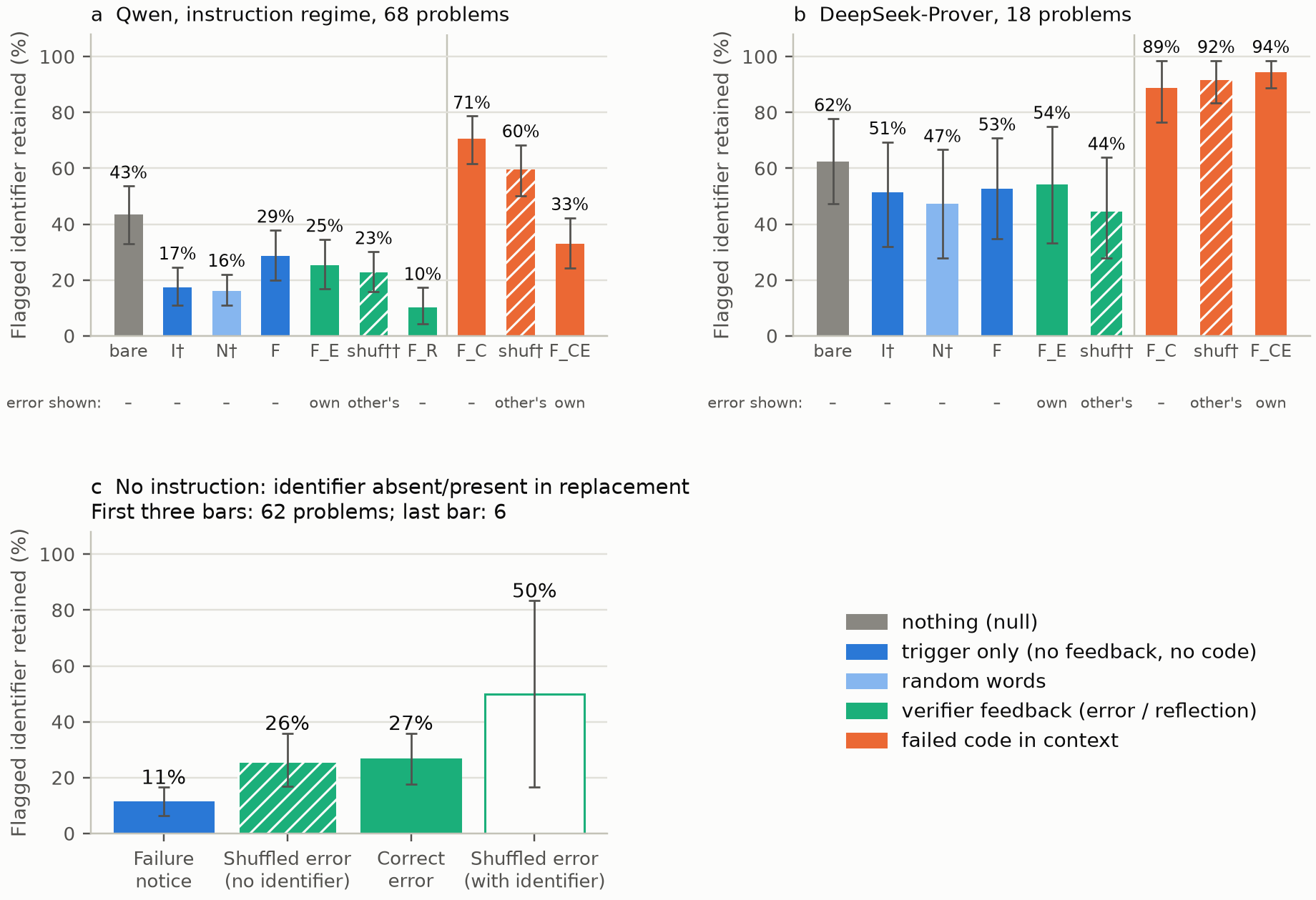}
\caption{Identifier retention. Top: Qwen and DeepSeek-Prover unknown-identifier subsets. Bottom: Qwen without the instruction; the first three bars use the 62 problems whose replacement diagnostic omits the identifier, and the last uses the six whose replacement includes it. Whiskers are 95\% problem-bootstrap intervals (2,000 resamples, seed 0). The panel heading is descriptive, not a causal interpretation. Copying and proof diversity appear in Fig.~\ref{fig:copy_details}.}
\label{fig:behavior_details}
\end{figure}

The stricter diagnostic-content control restricts the unknown-identifier subset to 62 problems whose replacement diagnostic does not contain the flagged identifier as a token. Without the instruction or failed proof, the identifier appears in 66/248 correct-diagnostic samples (26.6\%) and 64/248 shuffled-diagnostic samples (25.8\%; paired identifier-count sign $p=0.71$). Thus, its persistence cannot be explained solely by the replacement diagnostic repeating that identifier (Fig.~\ref{fig:behavior_details}, bottom). This is a behavioral check, not a test of equal proof success.

\begin{figure}[t]
\centering
\includegraphics[width=\linewidth]{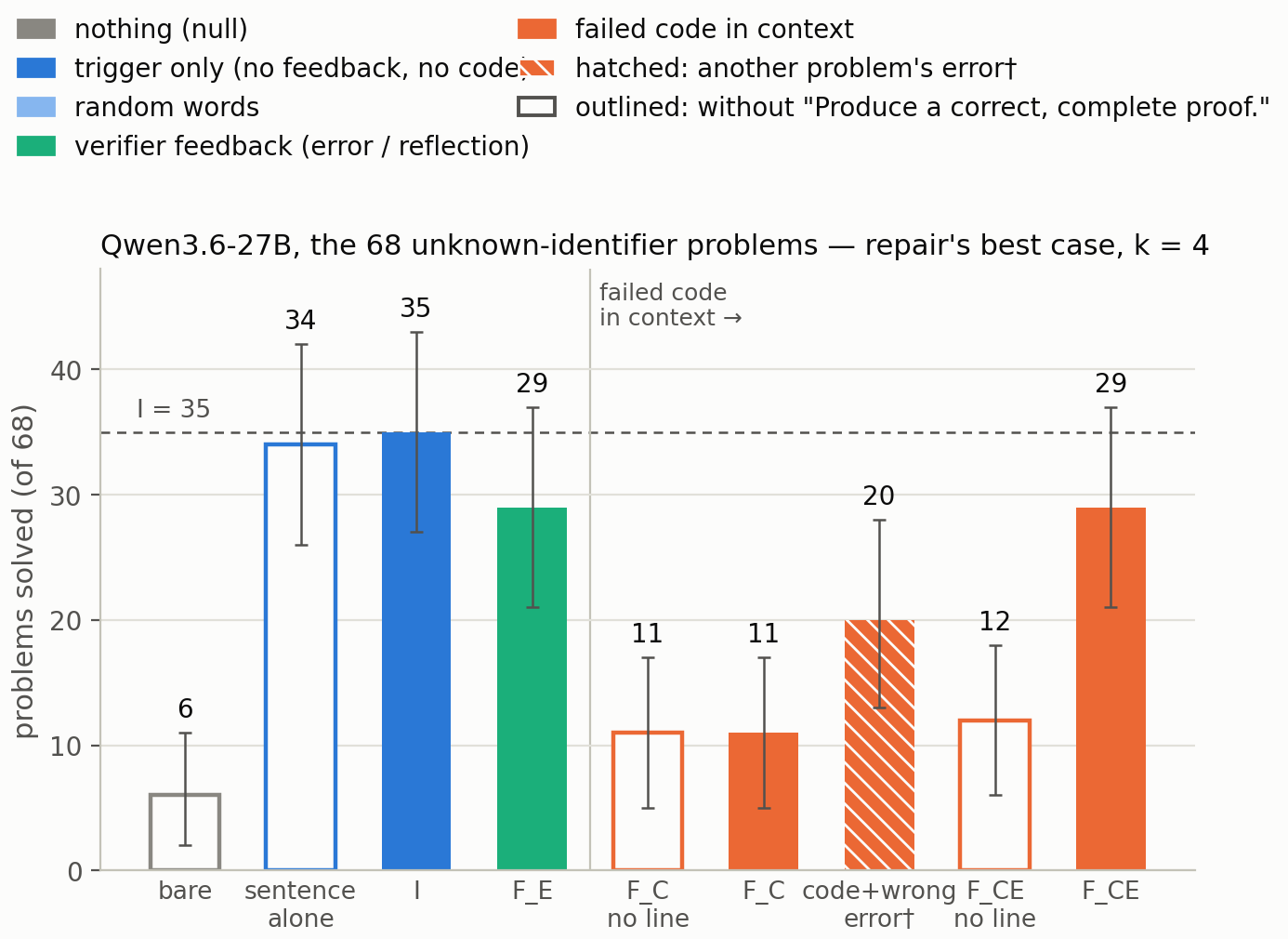}
\caption{Qwen unknown-identifier subgroup (68 problems). Correct diagnostics recover performance relative to code alone, but do not surpass instruction-only retrying. Outlined bars omit the instruction. The no-instruction code-plus-error comparison is unresolved; this subgroup accounts for 18 of the 22 problems recovered in the full bank (Table~\ref{tab:identifier-split}).}
\label{fig:subgroup_controls}
\end{figure}

The same subgroup analysis also reveals sensitivity to the instruction: among all 68 unknown-identifier problems, code plus correct diagnostics reaches 29 versus 11 with code alone when the instruction is present, but only 12 versus 11 without it (paired pass-count sign $p=0.58$; Fig.~\ref{fig:subgroup_controls}). The latter contrast does not resolve a recovery effect.

In the full Qwen bank, adding correct diagnostics to failed code reduces exact predecessor copies from 111 to 76 (35 fewer of 668 samples), while passing samples rise from 36 to 106 (70 more). These are separate aggregate changes, not a decomposition of the causal effect into copying and repair.

Copying is equality of extracted proof bodies after comment removal and whitespace normalization. Under $F_C$, own-predecessor copying is 16.6\% on Qwen and 60.8\% on DeepSeek-Prover; replacing the predecessor with another theorem's failed proof reduces those rates to 0.6\% and 1.6\%, and neither model copies the unrelated partner proof verbatim. With Qwen code-bearing diagnostics, copying is 11.4\% under the correct diagnostic, 23.7\% under the shuffled diagnostic, and 20.2\% under the class/length-matched wrong diagnostic. DeepSeek-Prover's correct- and shuffled-diagnostic rates are 53.8\% and 55.8\%. None of the copied own predecessors verifies. Mean distinct proofs among four samples fall from 3.77 under $F$ to 3.46 under $F_C$ on Qwen, and from 3.89 to 2.16 on DeepSeek. Qwen's correct-diagnostic condition has fewer exact predecessor copies but only 3.34 distinct proofs, and it does not reduce near-copying (see the approximate measures below): the diagnostic changes how the predecessor is reused more than whether it is reused.

On 68 Qwen problems, the predecessor diagnostic names an unknown identifier absent from the theorem statement. The identifier remains in 25.4\% of diagnostic-only samples, 22.8\% with shuffled diagnostics, and 10.3\% with reflection. Reflection thus changes a diagnostic-related behavior without surpassing instruction-only pass@4. With failed code present, identifier retention is 70.6\% for code alone, 59.6\% with shuffled diagnostics, and 33.1\% with correct diagnostics. The latter condition solves 29 of these problems, versus 11 with code alone and 20 with shuffled diagnostics; instruction-only solves 35. The remaining 99 problems contribute 11 versus 7 successes. Table~\ref{tab:identifier-split} compares the two subsets. Recovery and its advantage over the matched wrong diagnostic are concentrated in the identifier subset, with resolved differences between subsets; the failed-proof penalty is larger there, but its subset difference is not resolved. Even in this subset, $F_{CE}$ trails instruction-only retrying (29 versus 35). By predecessor length, Qwen's penalty is $-19.3$~pp ($[-31.6, -7.0]$) for the shortest third of predecessors and $0.0$~pp ($[-10.9, +10.9]$) for the longest; length and the identifier subset overlap, so we do not separate them. On DeepSeek-Prover's 18 identifier problems the recovery is $0.0$~pp, and its penalty lies in the other 119 problems ($-16.8$~pp; $[-23.5, -10.1]$). This subgroup was defined after the main experiment, so these comparisons are exploratory.

\begin{table}[t]
\centering
\caption{Qwen one-step contrasts split by whether the fixed predecessor's diagnostic names an unknown identifier absent from the theorem statement (exploratory). Entries are paired differences in percentage points with 95\% paired problem-bootstrap intervals (10{,}000 resamples); the last column resamples the two subsets independently. Parentheses give problems solved (pass@4) under the subtracted condition and then the first condition.}
\label{tab:identifier-split}
\resizebox{\linewidth}{!}{%
\begin{tabular}{lccc}
\toprule
Contrast & Identifier subset ($n=68$) & Other problems ($n=99$) & Subset difference \\
\midrule
$F_C - F$          & $-14.7$ $[-26.5, -2.9]$ ($21\to11$) & $-5.1$ $[-12.1, +2.0]$ ($12\to7$) & $-9.7$ $[-23.7, +4.6]$ \\
$F_{CE} - F_C$     & $+26.5$ $[+16.2, +38.2]$ ($11\to29$) & $+4.0$ $[-2.0, +10.1]$ ($7\to11$) & $+22.4$ $[+9.7, +35.3]$ \\
$F_{CE} - F_{CE_m}$ & $+17.6$ $[+7.4, +27.9]$ ($17\to29$) & $+1.0$ $[-5.1, +7.1]$ ($10\to11$) & $+16.6$ $[+4.3, +29.5]$ \\
$F_{CE} - I$       & $-8.8$ $[-19.1, +1.5]$ ($35\to29$) & $+1.0$ $[-5.1, +7.1]$ ($10\to11$) & --- \\
\bottomrule
\end{tabular}}
\end{table}

Without the instruction, identifier retention is 10.7\% for the failure notice alone and 28.3\% after adding the correct diagnostic. A shuffled diagnostic also increases retention (27.9\%). These measurements are consistent with sensitivity to feedback's presence and form, but do not establish that the model ignores its semantics. On DeepSeek's 18 unknown-identifier problems, retention is 54.2\% with diagnostic-only feedback versus 52.8\% under $F$; code-bearing conditions retain the identifier in 88.9--94.4\% of samples.

Proof similarity provides a further descriptive check: we compare normalized passing proof bodies with their predecessors using sequence similarity, truncated to the first 3,000 characters on each side. On Qwen's unknown-identifier subset, median similarity is 0.65 for code plus correct diagnostic (80 passing samples) versus 0.49 for instruction-only (86 passing samples); 31.3\% versus 12.8\% exceed 0.8 similarity. These passing-sample summaries are conditional on success and do not establish a causal editing mechanism. The proof examples appear in Appendix~\ref{app:examples}.

\paragraph{Toolchain drift and Lean~3 predecessors.}
\label{app:drift}
We classified every unknown identifier that a predecessor diagnostic names by checking it against the Mathlib, Batteries, and Lean core sources of our verifier (Mathlib v4.29.0) and of the environment DeepSeek-Prover-V2 was evaluated in (Lean 4.9, with the Mathlib pinned by DeepSeek-Prover-V1.5, committed in August 2024 with June 2024 content), with intermediate releases for deprecation records (Table~\ref{tab:drift}). All 18 of DeepSeek-Prover's flagged identifiers are valid names in its evaluation-era library that Mathlib has since renamed (for example \texttt{div\_le\_iff} to \texttt{div\_le\_iff}$_0$ and \texttt{add\_right\_neg} to \texttt{add\_neg\_cancel}). Of Qwen's 68, 48 are the Lean~3 keyword \texttt{begin}, 7 are renamed names, 1 needs an unopened namespace, 11 never existed, and 1 is a local hypothesis name. Across failed one-step samples, excluding those on statements the verifier cannot parse, unknown identifiers appear in 19\% of DeepSeek-Prover's failures, 93\% of them renamed names, and its sequential failures show the same pattern. DeepSeek-Prover's flat response to diagnostics (Section~\ref{sec:diagnostic_results}) may therefore partly reflect being told a lemma is unknown without knowing its new name.

Qwen's frozen bank contains 50 predecessors written in Lean~3 syntax, which a Lean~4 verifier rejects; 48 are the \texttt{begin} problems above. These are comparatively solvable problems (instruction-only retrying solves 31 of 50, against 14 of the other 117), and they carry most of Qwen's one-step success effects (Table~\ref{tab:lean3}): exposing a Lean~3 predecessor leads Qwen to write Lean~3 (73.5\% of $F_C$ samples contain \texttt{begin}, against 32.5\% under $F$), and the correct diagnostic undoes this. On the other 117 problems success is near the floor and the success effects are small and unresolved, although exact copying is as frequent (15.8\% of $F_C$ samples, against 18.5\% on Lean~3 predecessors). The sequential results do not share this composition: Lean~3 errors are under 1\% of failed attempts in the code-bearing histories, and among trajectories whose bare first attempt was not Lean~3, $CE$ solves 44.4\% against 20.5\% for $C$.

\begin{table}[h]
\centering
\caption{Unknown identifiers named by diagnostics, classified against the verifier's library and DeepSeek-Prover-V2's evaluation-era library. Predecessor columns count problems; sample columns count failed one-step samples on the one-step banks, excluding the runner's no-proof-body marker. Each diagnostic's first unknown identifier is resolved as Lean would under the proof's \texttt{open} declarations; single-letter names such as \texttt{I} and \texttt{X} that resolve under an opened namespace count as needing that namespace.}
\label{tab:drift}
\small
\begin{tabular}{lrrrr}
\toprule
 & \multicolumn{2}{c}{Predecessors} & \multicolumn{2}{c}{Failed one-step samples} \\
Category & Qwen & DeepSeek & Qwen & DeepSeek \\
\midrule
Lean~3 keyword (\texttt{begin}) & 48 & 0 & 2{,}376 & 0 \\
Valid in 2024 Mathlib, renamed since & 7 & 18 & 1{,}269 & 1{,}089 \\
Needs an unopened namespace & 1 & 0 & 49 & 0 \\
Never in either library & 11 & 0 & 1{,}135 & 20 \\
Local hypothesis or variable & 1 & 0 & 271 & 61 \\
Other & 0 & 0 & 104 & 7 \\
\midrule
Total & 68 & 18 & 5{,}204 & 1{,}177 \\
\bottomrule
\end{tabular}
\end{table}

\begin{table}[h]
\centering
\caption{Qwen one-step contrasts (pp) split by whether the frozen predecessor is written in Lean~3 syntax (exploratory; 10{,}000 problem-bootstrap resamples). Parentheses give problems solved under the subtracted and then the first condition.}
\label{tab:lean3}
\small
\begin{tabular}{lll}
\toprule
Contrast & Lean~3 predecessors ($n=50$) & Other problems ($n=117$) \\
\midrule
$F_C-F$ & $-26.0$ $[-40.0,-12.0]$ ($21\to8$) & $-1.7$ $[-8.5,+5.1]$ ($12\to10$) \\
$F_{CE}-F_C$ & $+34.0$ $[+20.0,+48.0]$ ($8\to25$) & $+4.3$ $[-0.9,+9.4]$ ($10\to15$) \\
$F_{CE}-F_{CE_m}$ & $+20.0$ $[+6.0,+34.0]$ ($15\to25$) & $+2.6$ $[-3.4,+8.5]$ ($12\to15$) \\
$F_{C_o}-F_C$ & $+42.0$ $[+26.0,+58.0]$ ($8\to29$) & $+2.6$ $[-3.4,+8.5]$ ($10\to13$) \\
$F_{CE}-I$ & $-12.0$ $[-26.0,0.0]$ ($31\to25$) & $+0.9$ $[-4.3,+6.0]$ ($14\to15$) \\
\bottomrule
\end{tabular}
\end{table}

\paragraph{Approximate copying measures.} Exact equality is interpretable but strict, so we recomputed copying over every retained sample on the full bank, without conditioning on success, adding three approximate measures: the share of samples within $0.9$ sequence similarity of the predecessor, mean sequence similarity, and mean Jaccard overlap of tactic-head multisets. Normalization and the 3,000-character truncation follow the measurement above. Exposing the own failed proof raises exact copying, near-copying, and mean similarity relative to the failure-notice scaffold, by $+15.4$ and $+58.2$~pp exact, $+21.0$ and $+65.7$~pp near-copying, and $+0.215$ and $+0.440$ in mean similarity on Qwen and DeepSeek-Prover, every interval excluding zero; predecessor replacement returns each measure to or below the scaffold level. The behavioral result therefore does not depend on exactness. One qualification concerns the diagnostic: on Qwen the correct diagnostic reduces exact copying relative to the failed proof alone ($-5.2$~pp; $[-9.7,-0.7]$) and tactic-head overlap ($-0.101$; $[-0.154,-0.049]$), but not near-copying ($-0.3$~pp; $[-5.8,+5.2]$) or mean similarity ($+0.005$; $[-0.030,+0.039]$), so it appears to change which tactics are tried while leaving the output about as close to the predecessor overall. On DeepSeek-Prover the reduction is resolved on all measures but small against a much higher base rate.

\paragraph{Many predecessors per theorem.} The one-step contrasts hold one failed predecessor fixed per theorem, so they cannot show whether exposure acts the same way on other failures of the same theorem. The sequential runs supply that variation without new generation: attempt 1 of every adaptive arm is a bare draw, so each trajectory that fails there and continues is a one-step repair on a freshly sampled predecessor. Comparing attempt 2 of the failed-proof history ($C$) against attempt 2 of the diagnostic-only history ($E$), conditional on attempt 1 failing, contrasts exposure and withholding across 107 and 109 Qwen predecessors and 103 and 101 on DeepSeek-Prover, paired at the theorem level and bootstrapped over theorems and then trajectories. Exposure reproduces its behavioral signature: mean similarity to the fresh attempt-1 proof rises by $+0.240$ on Qwen ($[+0.116,+0.368]$, positive on 9 of 11 theorems) and $+0.483$ on DeepSeek-Prover ($[+0.386,+0.575]$, positive on all 11), with exact reproduction at $+8.6$ and $+68.5$~pp. The success penalty is not resolved at a single repair step here, because attempt-2 success on these banks, averaged first within each theorem and then across theorems, is near the floor: $4.6$\% versus $2.8$\% on Qwen ($+1.8$~pp; $[-5.1,+10.2]$) and $2.7$\% versus $2.9$\% on DeepSeek-Prover ($-0.2$~pp; $[-7.2,+5.5]$). We therefore read this as evidence that the exposure-to-copying association generalizes across failure realizations, not as a multi-predecessor replication of the success penalty.

The union across Qwen conditions in the main, shuffled-diagnostic, code-plus-shuffled-diagnostic, and no-instruction runs covers 80/167 problems, versus 49 for the highest-count individual condition. This shows that the single-condition plateau is not a hard ceiling, but uses more generations and is not an equal-budget comparison.

\subsection{Sequential controls across models and robustness}
\label{app:sequential_results}

The sequential comparison evaluates Qwen and DeepSeek-Prover separately on their own 11-problem banks. Each model has ten trajectories per adaptive arm and a 64-attempt horizon, with early stopping at the first verified proof. Both analyses compare failed-proof history alone ($C$), with corresponding diagnostics ($CE$), ordinarily shuffled wrong diagnostics ($CE_s$), and class/length-matched wrong diagnostics ($CE_m$), together with code-free corresponding- and matched-wrong-diagnostic histories ($E$ and $E_m$). Artifact labels appear in Table~\ref{tab:aliases}. Memoryless $I$ is mean pooled pass@$k$ from 640 draws per problem. Matching controls and registration status are described in Appendix~\ref{sec:methods_sequential}. Figure~\ref{fig:sequential} shows both models, with Qwen above DeepSeek.

\paragraph{Preregistered Qwen repair-versus-retry endpoint.}
On Qwen at $k=64$, memoryless instruction-only retrying reaches mean pooled pass@$k$ of 62.8\%, whereas $CE$ reaches 49/110 trajectories (44.5\%). The preregistered primary contrast is therefore $CE-I=-18.3$ percentage points (pp), with a paired problem-bootstrap 95\% interval of $[-41.8,+5.3]$: the experiment does not establish an advantage for repair over retrying. The conclusion is unchanged when conditioning on failure of attempt 1 ($-19.1$ pp; $[-42.3,+4.5]$), which addresses the fact that adaptive arms begin with the bare completion prompt while memoryless $I$ receives the explicit instruction from its first draw. Code-only history is substantially worse than retrying: $C-I=-40.1$ pp ($[-62.2,-17.0]$).

\begin{table}[t]
\centering
\caption{Qwen sequential contrasts at $k=64$ on 11 problems, grouped by registration status. Primary and registered secondary (Reg.) contrasts were specified in a timestamped document before the corresponding arm was generated (Appendix~\ref{sec:methods_sequential}); exploratory (Expl.) contrasts were not. Intervals are unadjusted 95\% percentile intervals from 10{,}000 paired problem-bootstrap resamples that carry all trajectories and memoryless draws for each resampled problem. An interval spanning zero resolves neither a difference nor equivalence. The content-free filler arm (noise) exists only for Qwen, so $E_m-\text{noise}$ has no DeepSeek-Prover counterpart.}
\label{tab:qwen-seq}
\small
\begin{tabular}{llrl}
\toprule
Status & Contrast & $\Delta$ (pp) & 95\% CI \\
\midrule
\multicolumn{4}{l}{\emph{Ordinary-shuffle registration (2026-09-07, 2026-09-08)}} \\
Primary & $CE-I$ & $-18.3$ & $[-41.8,+5.3]$ \\
Primary & $(CE-CE_s)-(E-E_m)$ & $+11.8$ & $[-8.2,+29.1]$ \\
Reg. & $C-I$ & $-40.1$ & $[-62.2,-17.0]$ \\
Reg. & $CE-C$ & $+21.8$ & $[+10.9,+32.7]$ \\
Reg. & $CE-CE_s$ & $+18.2$ & $[+8.2,+28.2]$ \\
Reg. & $CE-E$ & $-19.1$ & $[-31.8,-5.5]$ \\
Reg. & $C-E$ & $-40.9$ & $[-54.5,-26.4]$ \\
Reg. & $CE_s-I$ & $-36.5$ & $[-58.4,-12.7]$ \\
Reg. & $E-E_m$ & $+6.4$ & $[-6.4,+19.1]$ \\
Reg. & $E_m-I$ & $-5.6$ & $[-26.2,+13.7]$ \\
Reg. & $E_m-\text{noise}$ & $-2.7$ & $[-16.4,+11.8]$ \\
Reg. & $CE_s-E_m$ & $-30.9$ & $[-49.1,-13.6]$ \\
\midrule
\multicolumn{4}{l}{\emph{Matched-diagnostic ($CE_m$) registration (2026-09-12)}} \\
Primary & $CE-CE_m$ & $+23.6$ & $[+12.7,+35.5]$ \\
Reg. & $(CE-CE_m)-(E-E_m)$ & $+17.3$ & $[-1.8,+34.5]$ \\
Reg. & $CE_m-I$ & $-41.9$ & $[-62.1,-22.2]$ \\
Reg. & $CE_m-C$ & $-1.8$ & $[-10.9,+9.1]$ \\
Reg. & $CE_m-CE_s$ & $-5.5$ & $[-17.3,+7.3]$ \\
\midrule
\multicolumn{4}{l}{\emph{Exploratory}} \\
Expl. & $E-I$ & $+0.8$ & $[-13.2,+16.4]$ \\
Expl. & $C-CE_s$ & $-3.6$ & $[-14.5,+7.3]$ \\
Expl. & $CE_s-E$ & $-37.3$ & $[-50.0,-25.5]$ \\
Expl. & $CE_m-E_m$ & $-36.4$ & $[-50.0,-21.8]$ \\
\bottomrule
\end{tabular}
\end{table}

\paragraph{Qwen diagnostic-content effect and preregistered interaction.}
Within Qwen code-bearing repair, corresponding diagnostic content has a large resolved effect: $CE-CE_s=+18.2$ pp ($[+8.2,+28.2]$). Without code, the matched comparison is smaller and unresolved: $E-E_m=+6.4$ pp ($[-6.4,+19.1]$), with 70/110 versus 63/110 trajectories solved by $k=64$. The preregistered difference-in-differences,
\[
(CE-CE_s)-(E-E_m),
\]
is $+11.8$ pp at $k=64$ with a 95\% interval of $[-8.2,+29.1]$ and therefore does not establish an interaction at the primary horizon. Its point estimate is positive across horizons and its interval is entirely above zero at several intermediate depths, but those secondary horizons do not replace the preregistered $k=64$ endpoint. We therefore report the component effects directly rather than inferring that diagnostic content matters \emph{only} when code is present.

The registered $CE_m$ arm supplies a symmetric matching-policy comparison. At $k=64$, $CE-CE_m=+23.6$ pp ($[+12.7,+35.5]$), while $(CE-CE_m)-(E-E_m)=+17.3$ pp ($[-1.8,+34.5]$). Thus Qwen's within-repair correspondence effect is resolved, but the matched interaction is not. The wrong-diagnostic policies themselves do not resolve a difference: $CE_m-CE_s=-5.5$ pp ($[-17.3,+7.3]$).

Qwen's $E_m$ control is closely matched to the corresponding diagnostic: among 4,055 wrong-error draws, 98.7\% use the same coarse error class, only 51 fall back to the pooled tactic family, and the wrong-to-own diagnostic token-length ratio is 1.00 at both quartiles. $E_m$ itself shows no resolved difference from content-free noise ($-2.7$ pp; $[-16.4,+11.8]$) and no advantage over memoryless retrying ($-5.6$ pp; $[-26.2,+13.7]$). Thus the code-free result is not explained by a poorly matched wrong-diagnostic control.

\paragraph{Qwen depth profile.}
Table~\ref{tab:seq_depth_final} reports the observed adaptive trajectories and the pooled memoryless estimate at the preregistered horizons. Qwen's $CE$ is close to the memoryless point estimate around $k=4$ and then accumulates success more slowly; $CE_s$ is below memoryless retrying from $k=8$ onward with intervals excluding zero in the preregistered analysis. The endpoint conclusions are based on paired problem-bootstrap contrasts, not on visual separation of marginal curves.

\begin{table}[h]\centering\small
\caption{Qwen sequential success by horizon. The $E$, noise, and assert rows come from the earlier August launch (Appendix~\ref{sec:methods_sequential}); assert shows only a failure marker per prior attempt. Adaptive entries are trajectories of 110 with a first verified pass by $k$; memoryless $I$ is the mean pooled pass@$k$ estimate from 640 independent draws per problem.}
\label{tab:seq_depth_final}
\begin{tabular}{lrrrrrrr}
\hline
arm & 1 & 2 & 4 & 8 & 16 & 32 & 64 \\ \hline
$C$ & 3 & 8 & 12 & 15 & 17 & 20 & 25 \\
$CE$ & 4 & 11 & 28 & 35 & 40 & 44 & 49 \\
$CE_m$ & 2 & 9 & 13 & 15 & 17 & 20 & 23 \\
$CE_s$ & 2 & 9 & 17 & 18 & 22 & 24 & 29 \\
$E$ & 1 & 4 & 14 & 24 & 40 & 60 & 70 \\
$E_m$ & 5 & 10 & 20 & 27 & 37 & 51 & 63 \\
noise & 1 & 7 & 15 & 30 & 42 & 60 & 66 \\
assert & 2 & 6 & 18 & 38 & 50 & 60 & 64 \\
$I$ (pooled) & 8.6\% & 15.0\% & 23.8\% & 34.4\% & 45.6\% & 55.8\% & 62.8\% \\
\hline
\end{tabular}
\end{table}

\paragraph{DeepSeek-Prover endpoints.}
At $k=64$, failed-proof history alone and with correct diagnostics each solve 27/110 trajectories (24.5\%), compared with 75.3\% for pooled memoryless $I$. Thus $C-I$ and $CE-I$ are both $-50.7$ pp, and both paired intervals exclude zero (Table~\ref{tab:seq_endpoints_deepseek}). Failed-proof history with wrong diagnostics is also below memoryless retrying. This supports failed-history harm in both tested banks: Qwen's $C-I$ and DeepSeek's $C-I$ are negative with intervals excluding zero. Qwen's $CE-I$ remains unresolved, whereas DeepSeek's $CE-I$ is entirely negative.

\begin{table}[t]
\centering
\caption{DeepSeek-Prover sequential contrasts at $k=64$ on its own 11-problem bank, grouped by registration status. The prover's sequential arms, including its ordinary-shuffle contrasts, were registered on 2026-09-11 before any prover depth attempt was generated. Entries are percentage-point differences with unadjusted 95\% percentile intervals from 200{,}000 paired problem-bootstrap resamples; the $CE_m$ family uses the registered 10{,}000 resamples. Memoryless $I$ uses pooled pass@$k$. $CE-E$ and $C-E$ coincide to the reported precision; their per-problem difference vectors differ, but with 11 problem clusters the bootstrap percentiles land on the same endpoints.}
\label{tab:seq_endpoints_deepseek}
\small
\begin{tabular}{llrl}
\toprule
Status & Contrast & $\Delta$ (pp) & 95\% CI \\
\midrule
\multicolumn{4}{l}{\emph{Ordinary-shuffle registration (2026-09-11)}} \\
Primary & $CE-I$ & $-50.7$ & $[-66.4,-34.5]$ \\
Reg. & $C-I$ & $-50.7$ & $[-67.6,-33.6]$ \\
Reg. & $CE-C$ & $0.0$ & $[-3.6,+3.6]$ \\
Reg. & $CE-CE_s$ & $+1.8$ & $[-10.0,+13.6]$ \\
Reg. & $CE-E$ & $-42.7$ & $[-61.8,-23.6]$ \\
Reg. & $C-E$ & $-42.7$ & $[-61.8,-23.6]$ \\
Reg. & $CE_s-I$ & $-52.6$ & $[-68.9,-35.6]$ \\
Reg. & $E-I$ & $-8.0$ & $[-26.2,+6.5]$ \\
Reg. & $E-E_m$ & $-14.5$ & $[-30.0,-2.7]$ \\
Reg. & $E_m-I$ & $+6.5$ & $[-3.4,+18.3]$ \\
Reg. & $CE_s-E_m$ & $-59.1$ & $[-76.4,-40.0]$ \\
Reg. & $(CE-CE_s)-(E-E_m)$ & $+16.4$ & $[-1.8,+34.5]$ \\
\midrule
\multicolumn{4}{l}{\emph{Matched-diagnostic ($CE_m$) registration (2026-09-12)}} \\
Primary & $CE-CE_m$ & $+1.8$ & $[-5.5,+9.1]$ \\
Reg. & $(CE-CE_m)-(E-E_m)$ & $+16.4$ & $[+0.9,+35.5]$ \\
Reg. & $CE_m-I$ & $-52.6$ & $[-66.4,-37.3]$ \\
Reg. & $CE_m-C$ & $-1.8$ & $[-10.0,+6.4]$ \\
Reg. & $CE_m-CE_s$ & $0.0$ & $[-12.7,+11.8]$ \\
\midrule
\multicolumn{4}{l}{\emph{Exploratory}} \\
Expl. & $C-CE_s$ & $+1.8$ & $[-10.0,+12.7]$ \\
Expl. & $CE_s-E$ & $-44.5$ & $[-64.5,-23.6]$ \\
Expl. & $CE_m-E_m$ & $-59.1$ & $[-71.8,-44.5]$ \\
\bottomrule
\end{tabular}
\end{table}

DeepSeek does not reproduce Qwen's resolved within-repair diagnostic recovery: $CE-C=0.0$ pp, $CE-CE_s=+1.8$ pp, and $CE-CE_m=+1.8$ pp, with all intervals spanning zero. Both wrong-diagnostic repair arms solve 25/110 trajectories. Without failed proofs, corresponding-diagnostic history solves 74/110 trajectories (67.3\%) versus 90/110 (81.8\%) for matched-wrong-diagnostic history; $E-E_m=-14.5$ pp with a bootstrap interval entirely below zero, although the exact randomization test below does not resolve it ($p=0.063$). This favors the matched-wrong-diagnostic arm on this bank, without establishing a general benefit of incorrect feedback. The symmetric matched interaction is $+16.4$ pp ($[+0.9,+35.5]$), but its sign comes from this negative code-free contrast. The exploratory matched-policy protocol contrast $CE_m-E_m$ is $-59.1$ pp ($[-71.8,-44.5]$). These comparisons involve evolving histories with different donor pools rather than a shared fixed predecessor.

\paragraph{DeepSeek-Prover depth profile.}
Table~\ref{tab:seq_depth_deepseek} reports all seven arms at the same seven horizons as the Qwen depth table. The adaptive entries are first-pass counts, whereas the memoryless entries use the pooled estimator. The four code-bearing histories remain near 23--25\% at $k=64$, while the code-free histories and memoryless resampling reach substantially higher endpoint success. Intermediate horizons describe the trajectories of these effects; they do not replace the reported $k=64$ comparisons.

\begin{table}[t]
\centering\small
\caption{DeepSeek-Prover sequential success by horizon. Adaptive entries are trajectories of 110 with a first verified pass by $k$; memoryless $I$ is mean pooled pass@$k$ from 640 draws per problem.}
\label{tab:seq_depth_deepseek}
\begin{tabular}{lrrrrrrr}
\hline
arm & 1 & 2 & 4 & 8 & 16 & 32 & 64 \\ \hline
$C$ & 7 & 10 & 11 & 14 & 16 & 20 & 27 \\
$CE$ & 9 & 9 & 12 & 14 & 16 & 20 & 27 \\
$CE_m$ & 11 & 11 & 13 & 14 & 15 & 17 & 25 \\
$CE_s$ & 9 & 9 & 9 & 11 & 14 & 21 & 25 \\
$E$ & 9 & 12 & 26 & 35 & 52 & 61 & 74 \\
$E_m$ & 9 & 17 & 24 & 35 & 53 & 72 & 90 \\
$I$ (pooled) & 8.1\% & 14.2\% & 22.9\% & 34.2\% & 48.2\% & 62.9\% & 75.3\% \\
\hline
\end{tabular}
\end{table}

\paragraph{Qwen within-trajectory repetition.}
The Qwen sequential runs reproduce the predecessor-repetition pattern. A byte-identical repeat of a prior proof body occurs in 73.9\% of $C$ attempts, 78.3\% of $CE_s$, and 65.7\% of $CE$, compared with 0.0\% for $E$, 0.4\% for $E_m$, and 0.2\% for assertion-only history. In the $CE_m$ arm, exact recurrence of any prior nonempty saved code occurs on 4,343/5,755 post-initial attempts (75.5\%) for Qwen and 4,792/5,793 (82.7\%) for DeepSeek. These rates are exploratory and survivor-weighted because longer unsuccessful trajectories contribute more attempts. A wrong diagnostic therefore does not break the repetition loop, whereas the corresponding diagnostic on Qwen lowers exact repetition relative to the code-bearing controls, though not near-repetition of the previous attempt (similarity at least 0.9 on 81.7\% of $CE$ versus 83.9\% of $C$ attempts), and substantially improves verified success. This is convergent behavioral evidence for predecessor anchoring, not a mediation analysis establishing that repetition causes the success loss.

\begin{figure}[t]
\centering
\includegraphics[width=\linewidth]{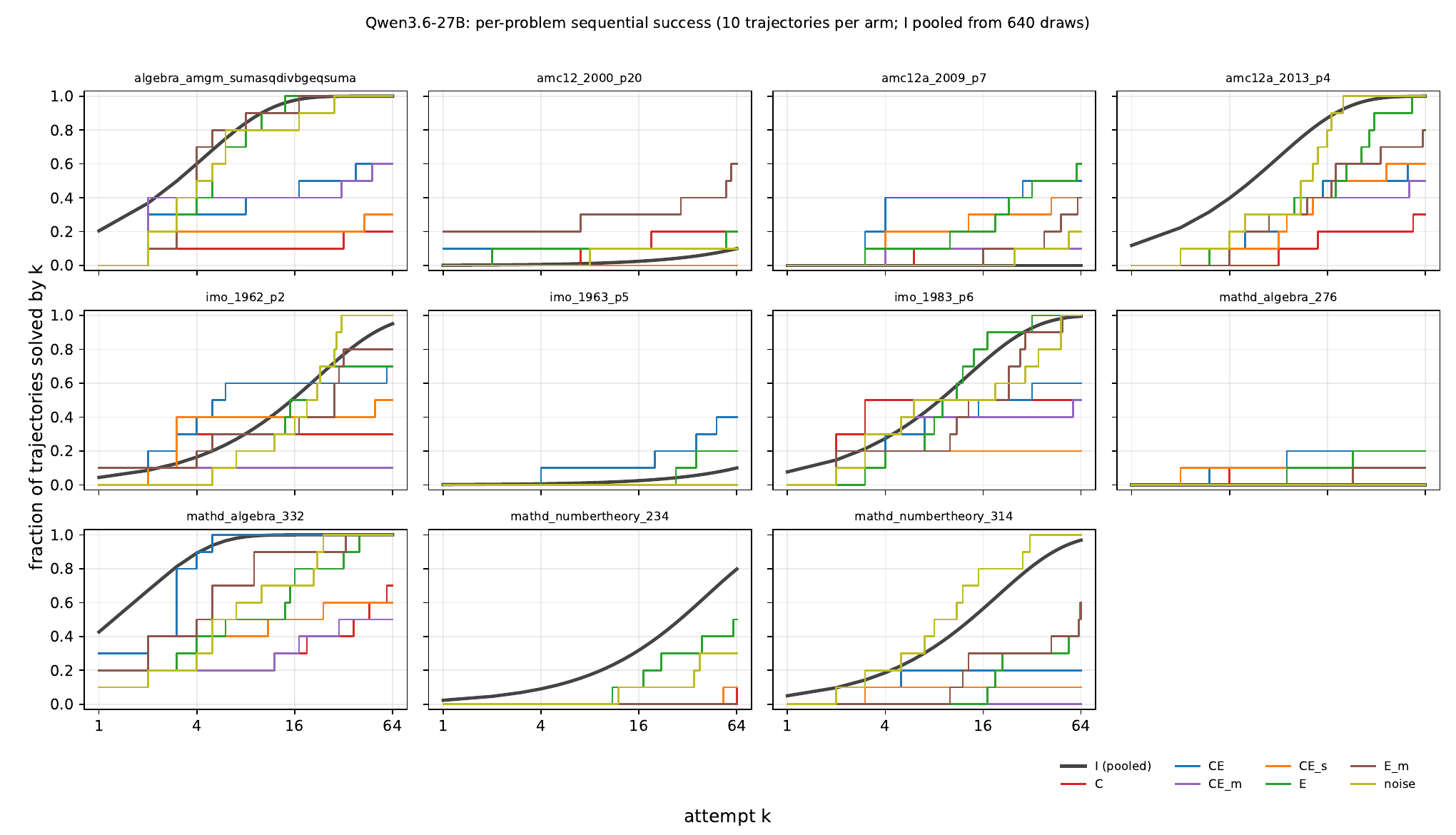}\\[4pt]
\includegraphics[width=\linewidth]{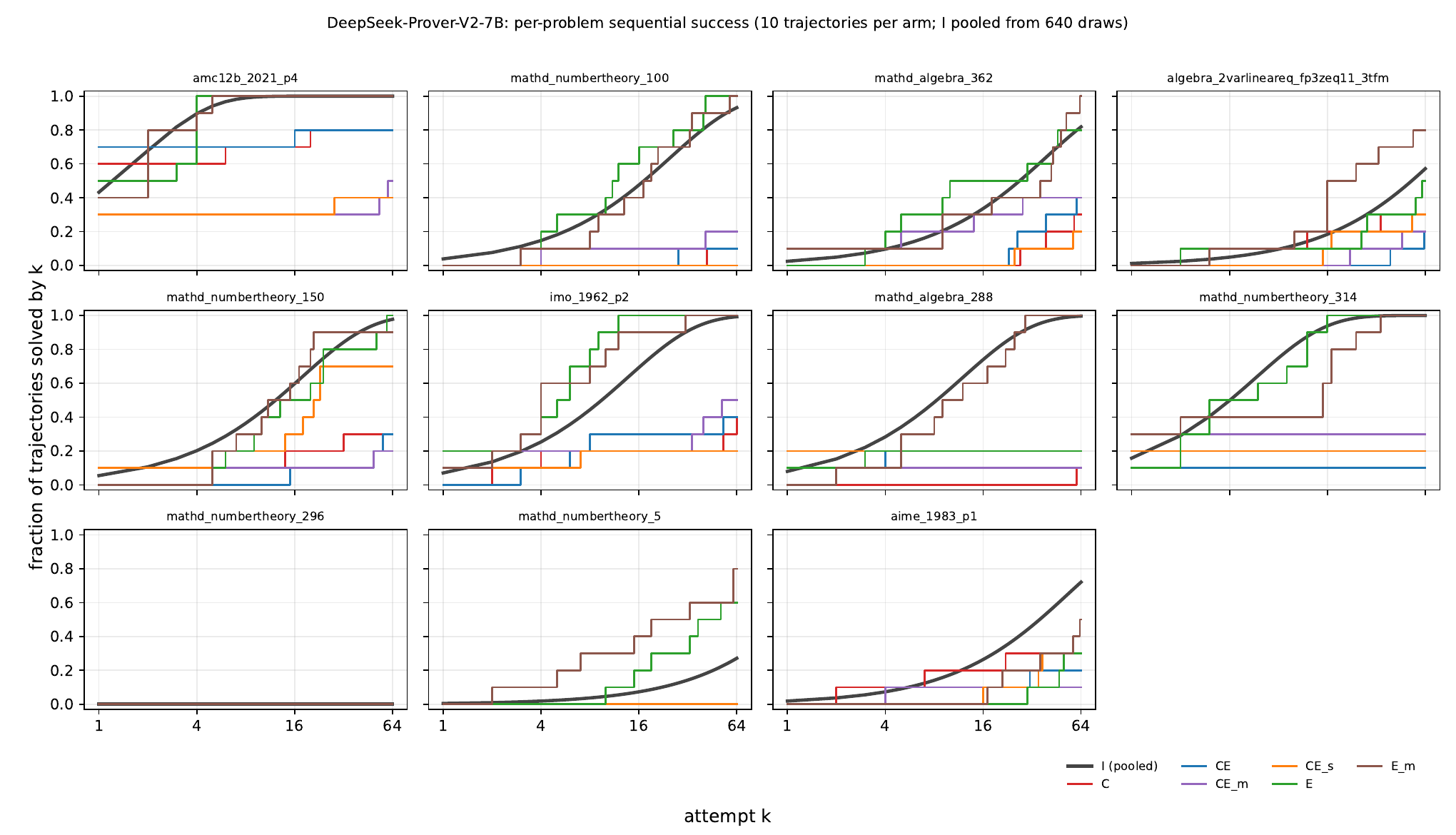}
\caption{Per-problem sequential success for Qwen (top) and DeepSeek-Prover (bottom). Adaptive curves are the fraction of ten trajectories solved by attempt $k$; memoryless $I$ is the pooled pass@$k$ from 640 draws on that problem.}
\label{fig:per_problem}
\end{figure}

\paragraph{Problem heterogeneity and uncertainty.}
Sequential inference uses only 11 problems per model and has correspondingly wide uncertainty; Fig.~\ref{fig:per_problem} shows every problem's curves. On Qwen, memoryless $I$ is near-certain to solve several problems by 64 draws and near-zero on several others; $CE$ has isolated problem-specific gains, including \texttt{imo\_1963\_p5}, but trails memoryless retrying on several problems that resampling solves reliably. Qwen's negative $CE-I$ point estimate therefore remains an unresolved difference, whereas DeepSeek's negative $CE-I$ interval supports underperformance on its bank. The banks overlap on only two problems and are not a paired cross-model experiment. A resolved diagnostic recovery on Qwen and an unresolved recovery contrast on DeepSeek do not by themselves establish a causal effect of model choice.

\paragraph{Leave-one-problem-out sensitivity.} With 11 problem clusters a single theorem can carry an endpoint, so we recomputed the main $k=64$ endpoints on each of the 11 leave-one-out banks (Table~\ref{tab:loo}). None of those that the paper reports as resolved changes sign on any leave-one-out bank. On Qwen, $CE-I$ stays within $[-25.1,-12.1]$, $C-I$ within $[-45.1,-34.4]$, $CE-C$ within $[+20.0,+25.0]$, $CE-CE_s$ within $[+16.0,+21.0]$, $CE-CE_m$ within $[+20.0,+26.0]$, and $CE_m-I$ within $[-47.1,-36.4]$. On DeepSeek-Prover, $CE-I$ stays within $[-55.8,-46.8]$, $E-E_m$ within $[-17.0,-8.0]$, and $CE_m-E_m$ within $[-65.0,-56.0]$. Sign changes occur only in contrasts the paper already declines to resolve: Qwen $E-I$ and $E_m-I$, and DeepSeek $CE-C$, $CE-CE_s$, and $CE-CE_m$, each of whose full-bank intervals spans zero. DeepSeek-Prover's matched interaction keeps its sign on every leave-one-out bank ($+9.0$ to $+20.0$~pp), but its interval no longer excludes zero on 6 of the 11, which is consistent with the exact test below. The leave-one-out range is therefore consistent with the reported uncertainty, and no resolved conclusion rests on one theorem.

\begin{table}[h]
\centering
\caption{Leave-one-problem-out sensitivity at $k=64$. ``Range'' is the smallest and largest point estimate over the 11 banks that each drop one problem. ``Sign'' counts banks whose estimate lies strictly on the other side of zero from the full-bank estimate, a zero full-bank estimate counting as nonnegative; ``Excl.'' counts banks on which the 95\% interval's exclusion of zero differs from the full bank's. Reduced-bank intervals use Python \texttt{random.Random(1)} with 10{,}000 resamples; full-bank intervals use \texttt{random.Random(0)} with 10{,}000 resamples on Qwen and the registered NumPy procedure on DeepSeek-Prover. The Qwen stream differs from the registered analysis, so its full-bank bounds can differ slightly from Table~\ref{tab:qwen-seq} (for $CE-I$, $[-42.1,+5.5]$ versus $[-41.8,+5.3]$).}
\label{tab:loo}
\small
\begin{tabular}{llrrrr}
\toprule
Model & Contrast & Full $\Delta$ (pp) & Range & Sign & Excl. \\
\midrule
Qwen & $CE-I$ & $-18.3$ & $[-25.1,-12.1]$ & 0 & 1 \\
 & $C-I$ & $-40.1$ & $[-45.1,-34.4]$ & 0 & 0 \\
 & $CE-C$ & $+21.8$ & $[+20.0,+25.0]$ & 0 & 0 \\
 & $CE-CE_s$ & $+18.2$ & $[+16.0,+21.0]$ & 0 & 0 \\
 & $CE-CE_m$ & $+23.6$ & $[+20.0,+26.0]$ & 0 & 0 \\
 & $CE_m-I$ & $-41.9$ & $[-47.1,-36.4]$ & 0 & 0 \\
 & $E-E_m$ & $+6.4$ & $[+2.0,+11.0]$ & 0 & 1 \\
 & $E-I$ & $+0.8$ & $[-5.1,+4.6]$ & 4 & 0 \\
 & $E_m-I$ & $-5.6$ & $[-11.1,+1.9]$ & 1 & 0 \\
 & $(CE-CE_s)-(E-E_m)$ & $+11.8$ & $[+8.0,+19.0]$ & 0 & 1 \\
 & $(CE-CE_m)-(E-E_m)$ & $+17.3$ & $[+12.0,+24.0]$ & 0 & 1 \\
\midrule
DeepSeek & $CE-I$ & $-50.7$ & $[-55.8,-46.8]$ & 0 & 0 \\
 & $C-I$ & $-50.7$ & $[-55.8,-46.8]$ & 0 & 0 \\
 & $CE-C$ & $0.0$ & $[-1.0,+1.0]$ & 2 & 0 \\
 & $CE-CE_s$ & $+1.8$ & $[-2.0,+6.0]$ & 1 & 0 \\
 & $CE-CE_m$ & $+1.8$ & $[-1.0,+4.0]$ & 1 & 0 \\
 & $E-E_m$ & $-14.5$ & $[-17.0,-8.0]$ & 0 & 0 \\
 & $CE_m-E_m$ & $-59.1$ & $[-65.0,-56.0]$ & 0 & 0 \\
 & $(CE-CE_s)-(E-E_m)$ & $+16.4$ & $[+10.0,+23.0]$ & 0 & 1 \\
 & $(CE-CE_m)-(E-E_m)$ & $+16.4$ & $[+9.0,+20.0]$ & 0 & 6 \\
\bottomrule
\end{tabular}
\end{table}

For Qwen's $CE-I$, $E-E_m$, and both interactions, and for DeepSeek-Prover's ordinary-shuffle interaction, the single change is a bank on which the interval newly excludes zero in the full-bank direction, not one that loses a resolved result.

\paragraph{Exact paired randomization tests.} Percentile bootstrap intervals over 11 problem clusters are coarse, so we also computed exact paired randomization tests: for each contrast we enumerated all $2^{11}=2048$ sign assignments of the per-problem differences at $k=64$ and took the two-sided proportion of assignments whose absolute mean difference is at least the observed absolute mean. This requires no resampling approximation. For the endpoints we tested, the two procedures agree on each resolved Qwen endpoint ($C-I$ $p=0.016$; $CE-C$ $p=0.012$; $CE-CE_s$ $p=0.012$; $CE-CE_m$ $p=0.004$; $CE_m-I$ $p=0.008$; $CE_s-E_m$ $p=0.016$; $CE_m-E_m$ $p=0.004$) and on those it does not ($CE-I$ $p=0.188$; $E-E_m$ $p=0.453$; both interactions $p\geq0.129$). They also agree on DeepSeek-Prover's large effects ($CE-I$, $C-I$, $CE_s-I$, $CE_m-I$, $CE_s-E_m$, $CE_m-E_m$, each $p=0.002$) and on the absence of within-repair recovery ($CE-C$ $p=1.000$; $CE-CE_s$ $p=0.883$; $CE-CE_m$ $p=0.828$). They disagree on two DeepSeek-Prover contrasts whose registered bootstrap intervals exclude zero: the code-free contrast $E-E_m$ ($p=0.063$, with only 6 of 11 problems contributing a nonzero difference) and the matched interaction whose sign it drives ($p=0.109$). We report the registered bootstrap as primary throughout and record the exact test as a sensitivity: it suggests that little weight should rest on DeepSeek-Prover's code-free contrast in particular. Either way the paper's conclusions do not rest on a DeepSeek-Prover diagnostic effect in either direction.

\paragraph{Launch provenance and a replicate memoryless baseline.} The $E$, noise, and assert rows of Table~\ref{tab:seq_depth_final} come from an earlier launch (27--29 August, tensor parallelism 4) than the other Qwen arms. That launch also drew 640 instruction-only memoryless draws per problem with the same prompt and seed grid as the registered memoryless $I$, but with tensor parallelism 4; we re-collected memoryless $I$ under its registration, with full per-draw logging and in the launch configuration of the code-bearing arms, and treat the earlier pool as a replicate. Only 1{,}173 of the 7{,}040 proofs are identical across the two launches. The pools agree at $k=1$ (8.5\% and 8.6\%) but differ at $k=64$ by 4.2~pp (67.09\% for the earlier pool versus 62.84\%), a gap driven by three problems with at most three successes in 640 draws, where a single success moves that problem's pooled pass@64 by about 10~pp. Against the earlier pool, $CE-I$ would be $-22.5$~pp ($[-44.4,+0.4]$) rather than $-18.3$~pp, and $E-I$ was $-3.45$~pp ($[-16.11,+10.89]$) rather than $+0.8$~pp, so the registered pool is the more favorable baseline for repair. Our intervals pool memoryless draws within a problem without resampling them; adding that second level widens our recomputation of Qwen's $CE-I$ interval from $[-42.1,+5.5]$ (a different resampling stream from the registered $[-41.8,+5.3]$) to $[-43.2,+7.0]$ and changes no conclusion. Contrasts involving $E$ or noise also compare runs from different launches, so their small differences, such as $E-I$, should be read against this replicate variation.

\subsection{Token expenditure}
\label{app:token_results}
\begin{figure}[t]
\centering
\includegraphics[width=\linewidth]{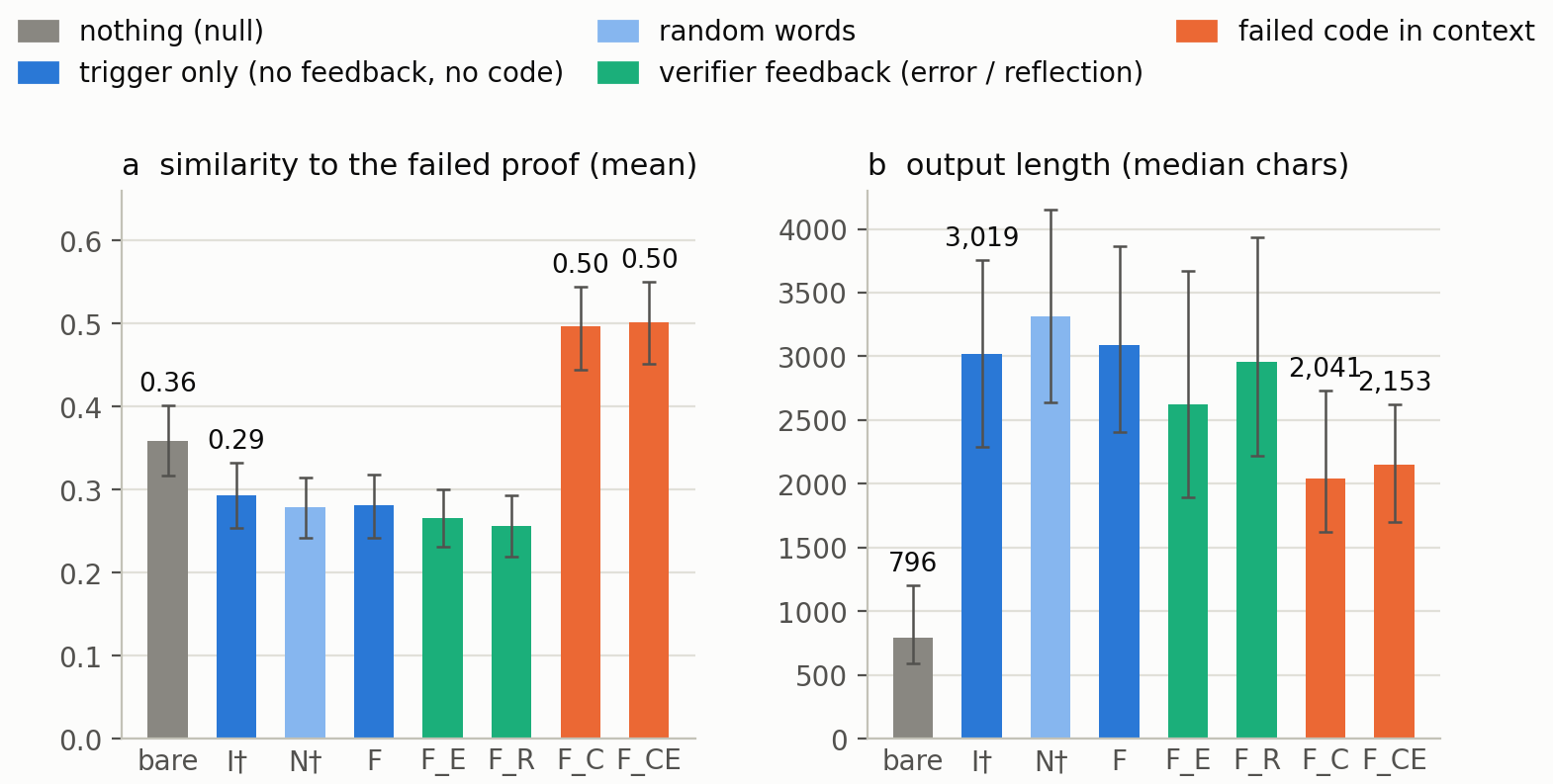}
\caption{Qwen all-sample similarity and visible output length. Similarity averages over parseable normalized proof bodies, comparing the first 3,000 characters with the predecessor. Output length is the median number of characters in retained response text. Whiskers use a 2,000-resample problem bootstrap (seed 0); they are not token-cost intervals. Daggers mark the figure's instruction-only and random-word post-hoc arms.}
\label{fig:similarity_length}
\end{figure}

Figure~\ref{fig:similarity_length} reports the all-sample similarity and output-length analysis. Mean predecessor similarity is approximately 0.50 with failed code (with or without the correct diagnostic), versus 0.29 under instruction-only prompting. This is distinct from the success-conditioned unknown-identifier analysis above. Median visible output lengths are 796 characters for bare, 3,019 for instruction-only, 2,041 for code alone, and 2,153 for code plus diagnostics. Character length describes the retained output text; the tokenizer-based measurements below are the relevant quantities for token expenditure.

For the Qwen one-step conditions, reconstructed mean input tokens per proof generation are 120 for $I$, 1,006 for $F_E$, 1,550 for $F_{CE}$, and 2,123 for $F_{CE_s}$. Medians are 113, 189, 606, and 642: long diagnostic outliers substantially affect means. Mean visible output tokens are 2,476, 2,479, 2,086, and 1,841, respectively. Reflection additionally requires a separate request, averaging approximately 1,565 input and 200 output tokens once per problem, shared across four proof continuations.

These counts retokenize retained visible responses and reconstructed prompts; they are not recorded billing or GPU compute. They exclude bank creation, verifier work, and infrastructure retries. The sequential comparison below matches budgets by token counts; it is not a FLOP- or latency-matched efficiency ranking.

\paragraph{Token-matched sequential comparison.} Our primary budget counts proof attempts, which equalizes verifier calls but not generation cost. Per-attempt token counts come from the full sequential traces, which the compact release omits; a per-attempt token table derived from them is released with the analysis: prompt and response tokens are runner-logged with the model's tokenizer for memoryless $I$ and for $C$, $CE$, $CE_s$, $CE_m$, and $E_m$. For Qwen's $E$, noise, and assert arms only compact records were retained, so their prompts are re-rendered with the runners' own history and eviction rules (the same rendering reproduces the logged $CE$ prompt counts exactly on 1{,}500 attempts), and their output tokens are predicted from the extracted proof by a linear fit on the logged $E_m$ rows, whose scaffold is identical to $E$'s ($R^2=0.43$; using $E_m$'s per-attempt mean instead changes $E$'s mean predicted output from 2{,}702 to 2{,}745 tokens). On DeepSeek-Prover, response tokens equal proof tokens, so every count is exact. Per attempt, Qwen's $CE$ uses 9{,}144 input and 1{,}640 output tokens on average, against 115 and 2{,}696 for a memoryless draw; DeepSeek-Prover's $CE$ uses 9{,}705 and 509, against 146 and 458.

Each protocol receives the same cap on cumulative tokens per trajectory and stops at its first verified proof; the memoryless curve averages 1{,}000 seeded random orderings of the 640 draws per problem. A trajectory that fails all 64 attempts is observed only up to what it spent, and the memoryless pool only up to its 640 draws, so we report a budget only below the smallest final spend of any failed trajectory in the arm and below the smallest 640-draw memoryless spend over problems. We do not compare memoryless success at an arm's mean cumulative cost: early stopping lets an adaptive arm concentrate spend on hard problems, which biases that comparison toward the arm. Intervals use 10{,}000 paired problem-bootstrap resamples, with exact randomization $p$-values. Table~\ref{tab:token_matched} reports the largest budgets at which every listed arm is observable under each accounting.

\begin{table}[t]
\centering
\caption{Sequential success at equal per-trajectory token budgets. Adaptive arms stop at their first verified proof and are capped at 64 attempts; memoryless $I$ draws until the budget is spent, so its number of draws depends on the budget (about 71 at Qwen's 200k total-token budget). Entries are arm minus memoryless $I$ in percentage points, with 95\% paired problem-bootstrap intervals and exact randomization $p$-values; the $I$ column is memoryless success at the same budget. Total and input accounting charge the retained history; output-only accounting is the most favorable to repair. Qwen's $E$ outputs are predicted (see text).}
\label{tab:token_matched}
\small
\setlength{\tabcolsep}{4pt}
\begin{tabular}{llrrlrlr}
\toprule
Model & Accounting & Budget & $I$ (\%) & $CE-I$ & $p$ & $E-I$ & $p$ \\
\midrule
Qwen & total & 200k & 63.1 & $-25.9$ $[-50.4,-2.4]$ & 0.066 & $-6.8$ $[-26.1,+10.9]$ & 0.50 \\
Qwen & input & 20k & 69.3 & $-40.2$ $[-64.4,-15.2]$ & 0.018 & $-26.6$ $[-44.5,-9.0]$ & 0.025 \\
Qwen & output & 20k & 35.9 & $-0.4$ $[-19.2,+17.7]$ & 0.98 & $-13.1$ $[-29.8,+2.7]$ & 0.16 \\
DeepSeek & total & 200k & 89.9 & $-73.5$ $[-89.8,-53.5]$ & 0.002 & $-26.3$ $[-46.3,-8.2]$ & 0.062 \\
DeepSeek & input & 50k & 90.3 & $-76.7$ $[-92.4,-56.7]$ & 0.002 & $-37.6$ $[-60.3,-16.1]$ & 0.016 \\
DeepSeek & output & 5k & 41.0 & $-28.2$ $[-43.0,-15.2]$ & 0.002 & $-1.0$ $[-9.3,+8.2]$ & 0.84 \\
\bottomrule
\end{tabular}
\end{table}

Under total or input accounting, repair falls further behind memoryless retrying than at equal attempts; only generated-token accounting brings Qwen's $CE$ level with it, and no accounting gives repair an advantage. Failed-proof history alone ($C$) trails further at every listed budget: $-45.9$, $-55.7$, and $-19.5$~pp on Qwen and $-71.7$, $-75.8$, and $-28.2$~pp on DeepSeek-Prover for total, input, and output accounting ($p\leq0.014$ except Qwen output, $p=0.14$). Per trajectory, through stopping or 64 attempts, Qwen's diagnostic-only history $E$ consumes about 15 times the input tokens and 1.3 times the total tokens of 64 memoryless draws, yet from 50{,}000 to 200{,}000 total tokens it stays within 9~pp of memoryless retrying without a resolved difference, while $CE$ trails by 12 to 26~pp over the same range; prompt length alone therefore does not explain the code-bearing deficit.

We do not report wall-clock or latency matching. The runs were generated on a shared cluster across several campaigns with differing concurrency, queueing, and serving configurations, so elapsed time is not a property of the conditions being compared. Verifier cost is equalized by construction in the attempt-matched comparison, since every arm issues one verification per attempt under the same verifier, timeout, and preamble; under token matching the memoryless arm issues proportionally more verifications, which makes its reported advantage conservative in verifier calls even as it becomes larger in generation tokens.

\section{Behavioral examples}
\label{app:examples}
Verbatim excerpts from the retained artifact rows (problem id, arm); proof bodies whitespace-normalized for display. Artifact access is described in Appendix~\ref{app:compute}.

\paragraph{1. Verbatim copying under the code arm} (\texttt{mathd\_numbertheory\_293}, \texttt{F\_C}). Shown its failed proof and told it did not verify, the model re-emits it byte-identically:
\begin{quote}\small\ttfamily
by norm\_num at h\textsubscript{1}; have h\textsubscript{2} : n $\leq$ 9 := h\textsubscript{0}; have h\textsubscript{3} : n $\geq$ 0 := by linarith; interval\_cases n <;> norm\_num at h\textsubscript{1}
\end{quote}
identical to the predecessor; the copy fails exactly as the predecessor did.

\paragraph{2. A local syntax repair} (\texttt{mathd\_algebra\_329}, \texttt{F\_CE}, PASS). The diagnostic flags \texttt{begin}, \texttt{have}, and \texttt{end} in a Lean~3-style predecessor. The passing proof retains the sequence of intermediate claims while replacing the block syntax and comma separators with Lean~4 syntax. This is consistent with local editing; the example alone does not identify how the diagnostic affected generation.
The predecessor starts with \texttt{begin have h\textsubscript{2} : x = 3 * y := by linarith, rw [h\textsubscript{2}] at h\textsubscript{1}, ...\ end}; the passing sample uses \texttt{by have h\textsubscript{2} : x = 3 * y := by linarith; rw [h\textsubscript{2}] at h\textsubscript{1}; ...}. The repair changes multiple syntax elements, rather than a single token.

\paragraph{3. Identifier reuse under an unrelated diagnostic} (\texttt{aime\_1997\_p9}, \texttt{F\_E\_shuf\_noI}). The replacement diagnostic is a type-mismatch report that never mentions \texttt{begin}; the sample nevertheless starts with \texttt{begin have h\textsubscript{4} : Int.floor (a\^{}2) = 2 := ...}. Appendix~\ref{app:behavior_results} tests identifier retention across the relevant subgroup rather than inferring a mechanism from this example.

\paragraph{4. Syntax correction is not sufficient for success} (\texttt{algebra\_2varlineareq\ldots}, \texttt{F\_R} versus \texttt{F\_E}). The reflection identifies the predecessor's \texttt{begin}/\texttt{end} syntax mismatch. All four reflection-conditioned samples avoid \texttt{begin}, but none verifies. Diagnostic-only prompting produces mixed outcomes: sample 0 retains \texttt{begin} and fails; samples 1 and 3 avoid it and verify; sample 2 avoids it but fails. Avoiding the obsolete syntax is therefore not sufficient for a verified proof.

\paragraph{5. The prover reuses the lemma its error names} (\texttt{mathd\_algebra\_171}, prover \texttt{F\_CE}). The error is \texttt{Unknown identifier `add\_right\_neg'}; the sample's \texttt{simp\_all} list includes \texttt{add\_right\_neg} anyway, and fails. With the code beside it, 94\% of the prover's samples on these problems still use the flagged lemma.

\paragraph{6. Matching a predecessor without seeing it.} The bare Qwen arm reproduces the normalized predecessor body on 7.2\% of samples, compared with 1.6\% under instruction-only retrying. Such matches can arise without proof exposure, so copying rates must be interpreted against a no-code baseline.

\section{Data integrity and re-verification}
\label{app:integrity}
The campaign encountered several distinct implementation issues. A serving configuration with Qwen thinking enabled was discarded; the analyzed run disables it. A preamble mismatch and toolchain purge required re-verification. A DeepSeek-Prover tokenizer/chat-template mismatch produced unparseable outputs; the analyzed replacement run includes a tokenizer guard. These are reasons to distinguish retained analysis artifacts from discarded launch outputs, not to pool them.

The corrected soundness screen recognizes Lean's backtick-quoted warning that a declaration uses \texttt{sorry}, including proofs written with \texttt{admit}. The retained re-verification audit records 3,197 rows classified as passing before correction, of which 224 became unsuccessful: 152 Qwen one-step rows, 13 Qwen extension rows, 11 adaptive-depth rows, and 48 memoryless rows from the earlier August launch (3 bare and 45 instruction-only; the latter belong to the replicate pool of Appendix~\ref{app:sequential_results}, not to the registered memoryless $I$); none of the re-verified DeepSeek rows flipped. The 11 adaptive trajectories stopped by invalid successes (one $E$, five noise, and five assert) were resumed with the invalidated attempt recorded as a failure whose diagnostic is Lean's warning that the declaration uses \texttt{sorry}; of these arms only $E$ displays diagnostics, so that warning entered one $E$ history as feedback. These counts describe the correction campaign rather than every file in the paper.

\paragraph{Complete re-verification of retained successes.} The 224 corrections group by condition as well as by file. All 224 are Qwen rows, and every one carried Lean's \texttt{sorry} warning rather than an unrelated error. They fall overwhelmingly in conditions that show the model no failed proof: 141 in the one-step content-free retry prompts (the bare, instruction-only, failure-notice, sentence-variant, and random-word arms), 59 in the sequential arms (45 of them in the replicate instruction-only memoryless pool), 10 in the one-step diagnostic arms that withhold the failed proof, 7 in reflection, and the remaining 7 in the only conditions that display the failed proof (6 $F_{CE}$ and 1 $F_C$). The uncorrected labels therefore inflated the instruction-only and memoryless baselines far more than the repair conditions, so correcting them moves the repair-versus-retry comparison slightly toward repair rather than away from it. Because the original correction script selected rows by a field name that the memoryless depth records do not use, its audit covered 3,197 of the then-passing rows rather than all of them. We therefore re-submitted every one of the 7,217 passing rows retained in the released artifact to the same Lean version under the corrected screen, across all 33 retained record files and both field conventions: all 7,217 were confirmed, none changed verdict, and no verifier failures occurred. The verdict log ships with the release.

The runners include completeness checks and duplicate-key checks, and the analysis scripts check the paired banks and sample counts. Prompt checks for the no-instruction and shuffled-diagnostic ablations establish the intended component changes. Such guards reduce implementation errors; they do not make all cross-protocol prompts identical.

The correction campaign uses retained corrected verdicts rather than generation-time success flags. Verification corrections included toolchain/preamble alignment and detection of proofs using \texttt{sorry} or \texttt{admit}. Infrastructure failures are not mathematical successes. The one-step replication retains two unresolved infrastructure rows as unsuccessful. Two of DeepSeek-Prover's 137 fixed one-step predecessors store a verifier-crash message rather than a compiler diagnostic; neither problem is solved under any one-step condition, so they contribute zero to every paired one-step contrast. In the sequential runners, transient verifier failures are retried three times; a failure that persists is recorded as an unsuccessful attempt under the prespecified rule, and its message then appears in later history as feedback. This affected one Qwen $E$ trajectory, which never solved (the crash message was displayed at attempts 14--64), and one DeepSeek-Prover $CE$ trajectory (displayed at attempts 26--46 before eviction; that trajectory solved at attempt 63). Each can move its arm's $k=64$ success by at most one trajectory in 110. The completed $E_m$ run (raw arm \texttt{E\_s}) contains one verifier-REPL infrastructure-failure row after retries and six truncated generations, all retained under that rule. Completeness and contiguous-attempt checks are run before endpoint analysis.

\paragraph{Seven-arm depth outcome audit.}
The Qwen and DeepSeek exports each cover 11 model-specific problems across the seven reported arms, with ten trajectories per problem in every adaptive arm and 640 memoryless draws per problem. Within each model, no duplicate $(\mathrm{problem},\mathrm{trajectory},\mathrm{attempt})$ keys, missing attempt indices, or unsuccessful trajectories ending before attempt 64 were found. Both schemas use attempt indices 1--64, including the memoryless \texttt{attempt\_id} field. Adaptive trajectories stop at their first retained \texttt{PASS} or reach attempt 64. The $CE_m$ mirrors contain 5,865 Qwen and 5,903 DeepSeek attempts; DeepSeek retains two verifier infrastructure failures as unsuccessful under the registered rule. Draw identity is determined by the attempt keys, not by the generation seed, which can repeat.

DeepSeek memoryless $I$ has 7,040 generated rows: 7,037 have an \texttt{ok} infrastructure status, and three on \texttt{mathd\_algebra\_288} report verifier failure after retries. Those three rows carry \texttt{FAIL} and remain in the 640-draw-per-problem denominator as unsuccessful attempts; no rows are dropped or relabeled. 

\section{Reproducibility and compute}
\label{app:compute}
\paragraph{Hardware and serving.}
The FarmShare generation campaign used SLURM-managed nodes equipped with NVIDIA L40S GPUs (approximately 46\,GB visible memory each). The Qwen sequential arms---memoryless $I$, $C$, $CE$, $CE_m$, $CE_s$, $E$, noise, and $E_m$---use the logged software environment vLLM 0.28.0, PyTorch 2.13, and Transformers 5.16.1 with the same verifier and decoding configuration. They were not generated in one launch: $E$, noise, and assert ran on 27--29 August 2026 with tensor parallelism 4, and memoryless $I$, $C$, $CE$, $CE_s$, $E_m$, and $CE_m$ on 6--12 September with tensor parallelism 2. Sampling with the same prompt and seeds is not reproducible across such launches (Appendix~\ref{app:sequential_results}). The memoryless control contains 7,040 draws (640 per problem); all generation and verification jobs completed, with two memoryless draws reaching the 16,384-token cap. The $CE_m$ run contains 5,865 retained rows across all 110 trajectories. The $E_m$ run (raw arm \texttt{E\_s}) contains 4,165 rows across all 110 trajectories, with one verifier-REPL infrastructure failure after retries and six truncated generations retained under the prespecified failure handling.

\paragraph{DeepSeek-Prover depth records.}
The DeepSeek depth export contains all seven analyzed arms on its 11-problem bank: 7,040 memoryless rows, 5,824 for $C$, 5,844 for $CE$, 5,903 for $CE_m$, 5,844 for $CE_s$, 3,457 for $E$ (raw \texttt{f}), and 2,857 for $E_m$ (raw \texttt{E\_s}). Adaptive row totals reflect stopping after the first retained pass. The $CE_m$ records contain two verifier infrastructure failures after retries, retained as unsuccessful attempts under the registered rule. Memoryless sampling records temperature 0.7 and a 16,384-token maximum; all 7,040 generations have \texttt{ok} generation status and \texttt{stop} finish reason. Three memoryless verifier failures after retries are retained as unsuccessful draws (Appendix~\ref{app:integrity}).

\paragraph{Partial GPU accounting.}
The available campaign accounting reports 171.9 productive GPU-hours, broken down in Table~\ref{tab:compute_partial}. This is a partial reported subtotal, not the total cost of all experiments included in the paper. The selection pool and depth runs outside that accounting are not covered, nor are complete costs established for the diagnostic replication or the Qwen and DeepSeek code-bearing sequential experiments. GPU-hours measure allocated GPUs multiplied by elapsed time, not accelerator utilization or model FLOPs.

\begin{table}[h]
\centering\small
\caption{Reported productive GPU-hours from the available campaign accounting. Categories follow that accounting and are not a complete inventory of the paper.}
\label{tab:compute_partial}
\begin{tabular}{lr}
\hline
Category & GPU-hours\\ \hline
Depth experiments & 124.8\\
Qwen one-step & 33.4\\
No-instruction ladder & 10.7\\
Code plus shuffled diagnostic & 1.7\\
DeepSeek-Prover one-step & 1.2\\
Probes & 0.1\\ \hline
Reported subtotal & 171.9\\ \hline
\end{tabular}
\end{table}
The same accounting separately reports 1.6 GPU-hours across the DeepSeek-Prover jobs it covers; this overlaps the categories above, must not be added to the subtotal, and is not a total for the DeepSeek depth runs.

\paragraph{CPU verification and software.}
The verification job scripts request 16 CPU cores and 48\,GB memory. The recorded operational estimate was approximately 50 minutes for 2,500 proofs (about 50 proofs per minute, or 13 allocated CPU-core-hours); this is a workload-specific estimate, not a uniform verifier throughput. Streaming verification overlapping generation is separate from the reported GPU accounting.

The recorded Qwen environment includes Python 3.12.3, PyTorch 2.13.0 with CUDA 13.0, Transformers 5.16.1, and FlashInfer 0.6.16.post3; the sequential runs log the same PyTorch and Transformers versions with vLLM 0.28.0. It records Kimina Lean Server v2.0.0, Lean v4.29.0, Mathlib revision \texttt{8a17838}, and REPL revision \texttt{4957772} for verification. These pins are campaign-specific and are not attributed to runs made under other environments.

\paragraph{Artifact labels.} Table~\ref{tab:aliases} maps the manuscript's arm names to the labels used in the released records.

\begin{table}[h]
\centering
\caption{Manuscript arm names and artifact labels. One-step labels are the \texttt{arm} field of the verified records; sequential labels name the files \texttt{depth\_<label>.jsonl}. The \texttt{\_noI} suffix marks a one-step arm without the retry instruction.}
\label{tab:aliases}
\small
\begin{tabular}{llll}
\toprule
\multicolumn{2}{l}{One-step} & \multicolumn{2}{l}{Sequential} \\
Manuscript & Artifact & Manuscript & Artifact \\
\midrule
bare, $I$, $F$ & \texttt{bare}, \texttt{I}, \texttt{F} & $I$ (memoryless) & \texttt{memI} \\
$F_C$, $F_E$, $F_{CE}$, $F_R$ & \texttt{F\_C}, \texttt{F\_E}, \texttt{F\_CE}, \texttt{F\_R} & $C$, $CE$ & \texttt{C}, \texttt{CE} \\
$F_{E_s}$, $F_{CE_s}$ & \texttt{F\_E\_shuf}, \texttt{F\_CE\_shuf} & $CE_s$, $CE_m$ & \texttt{CE\_s}, \texttt{CE\_m} \\
$F_{E_m}$, $F_{CE_m}$ & \texttt{F\_E\_match}, \texttt{F\_CE\_match} & $E$ & \texttt{f} \\
$F_{C_o}$ & \texttt{F\_C\_other} & $E_m$ & \texttt{E\_s} \\
random words, variants & \texttt{N}, \texttt{N2}--\texttt{N5}, \texttt{S}, \texttt{P}, \texttt{D} & noise, assert & \texttt{noise}, \texttt{assert} \\
no-instruction twins & \texttt{<arm>\_noI} & replicate memoryless pool & \texttt{blind\_I} \\
\bottomrule
\end{tabular}
\end{table}

\label{app:reproduce}
\paragraph{Reproduction artifacts.}
The recounts use frozen failure banks, retained per-sample records, and separate scripts for the main figures, pooled diagnostic comparison, benefit bounds, allocation estimates, and supporting checks. Regenerating these analyses does not require model generation or Lean execution. The two-model sequential analysis includes normalized problem-level curves, horizon counts, endpoint contrasts, source hashes, and completeness checks. Its reproducible scripts are \texttt{analyze\_sequential.py} and \texttt{plot\_sequential.py}; the former reads the seven-arm Qwen and DeepSeek archives, and the latter reads \texttt{per\_problem\_curves.csv} to regenerate the figure and every plotted confidence bound. The figure uses Qwen's 2,000-resample plotting procedure and the 200,000-resample DeepSeek procedure for the ordinary-shuffle arm family; the separately registered $CE_m$ endpoint family uses its documented 10,000-resample procedure (Appendix~\ref{app:stats}).

Code, the frozen failure banks, retained per-sample records, the re-verification verdict log, the registration documents, and every analysis script, including those for the analyses added in this version, are released at \url{https://github.com/mathingleft/repair-retry-hindsight} under the MIT license, with the records under CC BY 4.0.
Appendix~\ref{app:token_results} reports reconstructed token expenditure and its coverage limitations, separately from hardware allocation.
\section{Exploratory sentence variants}
\label{app:withdrawn}
The post-hoc sentence variants \texttt{S}, \texttt{P}, and \texttt{D} reach 30, 30, and 35 Qwen problems, compared with 33 for $F$ and 45 for $I$ (Table~\ref{tab:qwen}). These exploratory, unadjusted comparisons do not isolate an effect of sentence meaning: each variant uses the combined scaffold whose framing ablation is reported in Appendix~\ref{app:framing_results}. Additional random-word strings test sensitivity to the particular string, not a general benefit from arbitrary text.

\end{document}